\documentclass[lettersize,journal]{IEEEtran}
\usepackage{amsmath,amsfonts}
\usepackage{array}
\usepackage{subfigure}
\usepackage{textcomp}
\usepackage{stfloats}
\usepackage{url}
\usepackage{verbatim}
\usepackage{graphicx}
\usepackage{cite}
\usepackage{subfiles}
\usepackage{xcolor}
\usepackage{soul}
\usepackage{adjustbox}
\usepackage{algorithmic, algorithm}
\usepackage{epstopdf} 

\usepackage{booktabs}
\usepackage{multirow}

\renewcommand{\vec}[1]{\mbox{\boldmath${#1}$}}
\newcommand{\h}{\vec{h}}
\newcommand{\m}{^{(m)}}

\newcommand{\real}{\mathbb{R}}

\begin{document}

\bstctlcite{IEEEexample:BSTcontrol}

\title{Federated Learning in Mobile Networks:\\A Comprehensive Case Study on Traffic Forecasting}

\author{Nikolaos~Pavlidis, Vasileios~Perifanis, Selim~F.~Yilmaz, Francesc~Wilhelmi, Marco~Miozzo, Pavlos~S.~Efraimidis, Remous-Aris~Koutsiamanis, Pavol~Mulinka, Paolo~Dini
\thanks{N. Pavlidis, V. Perifanis and P. S. Efraimidis are with Department of Electrical and Computer Engineering, Democritus University of Thrace, Xanthi, Greece. S.\ F.\ Yilmaz is with Department of Electrical and Electronic Engineering, Imperial College London, London, United Kingdom. P  Mulinka, M. Miozzo and P. Dini are with Sustainable Artificial Intelligence, Centre Tecnol\`ogic de Telecomunicacions de Catalunya (CTTC/CERCA), Barcelona, Spain. F. Wilhelmi is with Nokia Bell Labs, Stuttgart, Germany. R.-A. Koutsiamanis is with Department of Automation, Production and Computer Sciences, IMT Atlantique, Inria, LS2N, Nantes, France.
}
}

\maketitle

\begin{abstract}
The increasing demand for efficient resource allocation in mobile networks has catalyzed the exploration of innovative solutions that could enhance the task of real-time cellular traffic prediction. Under these circumstances, federated learning (FL) stands out as a distributed and privacy-preserving solution to foster collaboration among different sites, thus enabling responsive near-the-edge solutions. In this paper, we comprehensively study the potential benefits of FL in telecommunications through a case study on federated traffic forecasting using real-world data from base stations (BSs) in Barcelona (Spain). Our study encompasses relevant aspects within the federated experience, including model aggregation techniques, outlier management, the impact of individual clients, personalized learning, and the integration of exogenous sources of data. The performed evaluation is based on both prediction accuracy and sustainability, thus showcasing the environmental impact of employed FL algorithms in various settings. The findings from our study highlight FL as a promising and robust solution for mobile traffic prediction, emphasizing its twin merits as a privacy-conscious and environmentally sustainable approach, while also demonstrating its capability to overcome data heterogeneity and ensure high-quality predictions, marking a significant stride towards its integration in mobile traffic management systems.
\end{abstract}


\section{Introduction}

Mobile traffic prediction is one of the popular topics for network optimization in fifth-generation (5G) systems and beyond. The ability to forecast mobile traffic patterns is crucial for operators to design and plan their networks efficiently, perform resource allocation, or mitigate anomalies (e.g., security threats), thereby enhancing the network’s quality of experience (QoE), resilience, and efficiency~\cite{shafiq2011characterizing}. 

In recent years, machine learning (ML) methods have become increasingly used to tackle traffic prediction in mobile networks~\cite{wang2017machine}, offering an attractive balance between the accuracy of the generated predictions and the timescale on which these are delivered when compared to alternative approaches such as analytical models or network simulators~\cite{ferriol2022building}. More specifically, deep learning (DL) has received significant attention due to its ability to capture the dynamics of network traffic both in terms of spatial and temporal dependencies across sites~\cite{lv2014traffic, yin2021deep}. DL models leverage network data measurements to predict future performance and utilization.

While DL can provide highly accurate predictions for network traffic forecasting~\cite{zhang2018long}, its use is often compute-intensive, particularly when it requires substantial training datasets to achieve the desired accuracy~\cite{cho2015much}. Indeed, DL relies upon deep architectures of artificial neural networks (ANNs) consisting of numerous neurons and layers, necessitating vast amounts of data to optimize the numerous parameters they encompass and deliver precise outcomes. This raises serious concerns regarding energy efficiency and environmental impact since, despite traffic forecasting can contribute to reducing energy consumption in mobile networks~\cite{piovesan2021mobile,piovesan2021forecasting}, the energy consumption needs of its corresponding DL forecasters remain largely unexplored and is anticipated to be significant.

To fully and cost-effectively leverage the potential of DL for mobile traffic prediction, we envision a scenario where multiple network sites collaboratively train an ML model on the edge, thus fitting in a multi-access edge computing (MEC) where network-related computations are made at the network edge~\cite{Ahvar2022}. In particular, we study the application of federated learning (FL)~\cite{mcmahan2017communication, konevcny2016federated} to traffic prediction. In a nutshell, FL is a distributed learning optimization framework that allows training ML models via the exchange of ML model updates rather than raw training data. The FL approach does not only reduce the overheads associated with data sharing in traditional centralized ML approaches but also safeguards data owners’ privacy. Furthermore, the privacy properties of FL allow the collaboration of multiple parties (e.g., network operators) to exchange insights on their proprietary datasets (typically kept private due to confidentiality constraints), in order to build even more robust and accurate collaborative models.

This article builds upon the work presented in \cite{perifanis2023federated}, which gathered the efforts from the winning team of the ``Federated traffic prediction for 5G and beyond'' problem statement in the ITU-T AI for 5G Challenge, and further delves into the application of FL for traffic forecasting in mobile networks. By using real measurements from the Long Term Evolution (LTE)’s Physical Downlink Control Channel (PDCCH) data, collected during multiple measurement campaigns at five base stations (BSs) across diverse and representative areas of Barcelona (Spain)~\cite{trinh2020mobile}, we highlight the suitability of FL models in terms of accuracy and energy efficiency, as well as on considerations for adoption. The specific contributions of this article are as follows:
 
 \begin{itemize}
    
    \item We introduce a comprehensive FL-based deep learning framework for mobile traffic forecasting along with a novel indicator to systematically measure the sustainability of the compared models. 
    
    \item We present a case study on federated traffic prediction using real data measurements from Barcelona (Spain). 
    
    \item We provide an exhaustive performance evaluation of the proposed solution, from both performance and sustainability perspectives. 
    
    \item We perform comprehensive ablation studies regarding key aspects of FL, including model aggregation strategies, the impact of individual clients, personalized learning, and utilization of exogenous data.     
    
\end{itemize}

The remainder of this document is organized as follows: Section~\ref{section:related_work} reviews the existing literature on time series forecasting for mobile traffic prediction and related FL solutions. Section~\ref{section:problem_formulation} introduces the problem formulation and our suggested FL-based methodology. Section~\ref{section:dataset} describes and analyzes the dataset used. Section~\ref{section:performance_evaluation} evaluates the proposed solution under various settings and conditions. Section~\ref{section:conclusions} summarizes the article with concluding remarks and potential future directions.

\section{Related Work}
\label{section:related_work}

\subsection{Time Series Forecasting}

The time series forecasting problem has been traditionally addressed through statistical-based models. In this regard, autoregressive (AR) and moving average (MA) methods like autoregressive integrated moving average (ARIMA) contributed to building the basics of time series prediction due to their ability to provide good short-term predictions~\cite{box1976time}. However, these methods fail at capturing seasonality and burstiness, thus lacking applicability to more complex time series forecasting problems (e.g., non-linear time series).

The complexity of non-linear time series problems can be properly addressed by sophisticated estimation methods like ANNs~\cite{hill1996neural}. Recurrent neural networks (RNNs) are a special type of ANN that is specifically designed for time series~\cite{mikolov2010recurrent}, thus making them useful in problems where sequential patterns are to be learned (e.g., speech recognition or video processing tasks~\cite{graves2013speech, guera2018deepfake}). However, RNNs fail at capturing seasonality, thus lacking applicability in mobile traffic forecasting. In particular, RNNs suffer gradient vanishing, as they are designed to capture infinite temporal relations. Nevertheless, a special type of RNN, namely long short-term memory (LSTM), has been widely used in communications, and more particularly, for mobile traffic forecasting~\cite{trinh2018mobile, trinh2019urban, qiu2018spatio}. LSTMs leverage \textit{forget gates} to avoid learning long-term trends in time series, thus allowing them to adjust the importance of old measurements onto future predictions. 

Another suitable neural network model for time series prediction is convolutional neural networks (CNNs), which has had tremendous success in image processing applications~\cite{gu2018recent}. CNNs come into play as the complexity of time series problems increases, including multi-variate problems or problems with multiple complex correlations between multiple types of features (e.g., spatial correlations, external phenomena, etc.). 
A CNN-based approach was introduced in~\cite{andreoletti2019network} for traffic prediction, which leveraged the topological structure of the network to generate accurate predictions.

More recently, complex DL models like ResNet~\cite{he2016deep} or VGG~\cite{simonyan2014very} were applied to problems in communications. For instance, in~\cite{han2021multivariate}, CNN, ResNet, and VGG were evaluated for multi-time series prediction in the Internet of things (IoT). However, while complex DL models provide outstanding results in fulfilling the time series prediction task, it comes at the expense of a very high computational cost. This was also confirmed in~\cite{perifanis2023towards}, where transformer architectures were evaluated in the context of federated traffic prediction, concluding that small accuracy improvements compared to simpler models such as LSTM or CNNs were achieved at a high cost in terms of energy consumption.

\subsection{FL for Time Series Forecasting}

While DL-based time series forecasting is a consolidated topic nowadays, distributed learning approaches for mobile traffic prediction remain highly unexplored. In this regard, FL was first applied in~\cite{liu2020privacy, zeng2021multi} for a similar problem, namely road traffic flow prediction. Different from centralized approaches (see, e.g.,~\cite{trinh2020mobile}), where a single model is generated for each location, the federated models are trained collaboratively, thus targeting robust collaborative models. The FL approach has been shown to leverage data from multiple users while preserving privacy, leading to similar predictive accuracy to centralized methods. 

Closer in spirit to our work, we find~\cite{zhang2021dual, zhang2022efficient}. First, baseline models such as support vector regression (SVR) and LSTM were trained federatively in~\cite{zhang2021dual}. To boost training efficiency, the authors proposed a clustering method whereby the contributions of the most significant BSs were accounted for. An extension of this work was proposed in~\cite{zhang2022efficient}, where distances between BSs were considered to model weights in the federated setting. Moreover, to increase accuracy at specific sites, personalized fine-tuning was proposed to be run on each BS (using local data) after training the global model. Both models in~\cite{zhang2021dual, zhang2022efficient} were trained and validated in a dataset taken by the mobile operator Telecom Italia in the areas of Milan and Trentino, between 2013 and 2014~\cite{barlacchi2015multi}. The dataset contains two months of data from SMS, voice calls, and Internet services records at up to 16.575 cells. Although the dataset in~\cite{barlacchi2015multi} has enabled the proliferation of traffic prediction models for communications in recent years, it is nowadays outdated since it does not capture novel use cases from current cellular networks. For that reason, in this work, we use a newer dataset that captures more recent user patterns in utilizing the network, such as massive multimedia services.

In our analysis, we focus on two FL issues that remain relatively unexplored in the field of cellular traffic forecasting, i.e., contribution of individual BS to the global model and combining exogenous data sources to improve accuracy and decrease energy costs.

In FL, handling heterogeneous clients is crucial for enhancing model performance and efficiency. The unique challenges presented by FL, including client data heterogeneity and variable computational capabilities, necessitate novel approaches for distributed optimization and privacy preservation. Li et al.~\cite{li2020federated} provided a comprehensive overview of current methodologies and future research directions within the FL landscape. Huang et al.~\cite{huang2023maverick} underscored the shortcomings of traditional contribution metrics like the Shapley Value in accurately assessing the contributions of Mavericks, clients with distinctive data distributions or exclusive data types. To address this, they introduced FedEMD, an adaptive client selection strategy leveraging the Wasserstein distance between local and global data distributions to enhance convergence and model accuracy by ensuring the preferential selection of Mavericks when beneficial. Xue et al.~\cite{xue2021toward} introduced the concept of Fed-Influence to quantify an individual client's impact on the federated model, facilitating a more nuanced approach to client selection and model debugging. These contributions collectively underscore the complexity of integrating diverse client inputs in FL and the necessity for innovative client selection strategies to optimize model performance.

Finally, using outer world data can boost the ML model's predictive accuracy for problems with strong spatio-temporal events. For instance, Essien et al.~\cite{essien2021deep} have shown that the fusion of traffic data with weather information can improve the accuracy of urban traffic prediction. This approach leverages a bi-directional LSTM network that includes weather data and other disruptive events reported on social media to predict traffic flows. In mobile traffic forecasting, He et al.~\cite{he2020graph} proposed the graph attention spatial-temporal network (GASTN), which captures local geographical dependencies and distant inter-region relationships through spatial and temporal data.

\section{Problem Formulation and Methodology}
\label{section:problem_formulation}

\subsection{Traffic Forecasting via FL}

We consider a multi-step time series forecasting problem on a cellular network with $K$ BSs connected to a common edge or cloud server. At every timestep $t$, each BS $k$ obtains a measurement vector $x_t^{(k)} \in \mathbb{R}^d$.  
Then, using previous samples $x_{1}^{(k)},\cdots,x_{t-1}^{(k)}$ , each BS $k$ predicts target measurement vector $Y_{t+T-1}^{(k)} \in \mathbb{R}^{T \times d'}$, where $d' \subseteq d$ denotes the set of features to be predicted and $T$ the number of steps predicted in the future. 

We utilize a common neural network model $f(\cdot)$ to make prediction, i.e., $\hat{y}_T^{(k)} = f( x_{1}^{(k)},\cdots,x_{t-1}^{(k)} )$, which is trained towards minimizing the mean squared error (MSE) of the prediction:

\begin{equation}
\label{eq:mse}
    \mathrm{MSE} = \frac{\sum_{t=1}^{T}\left(\sum_{i=1}^{d'} \left(\hat{y}^{(k)}_{t,i} - y^{(k)}_{t,i} \right)^2\right)}{Td'}.
\end{equation}

\begin{figure}[ht!]
    \centering    
    \includegraphics[width=\columnwidth]{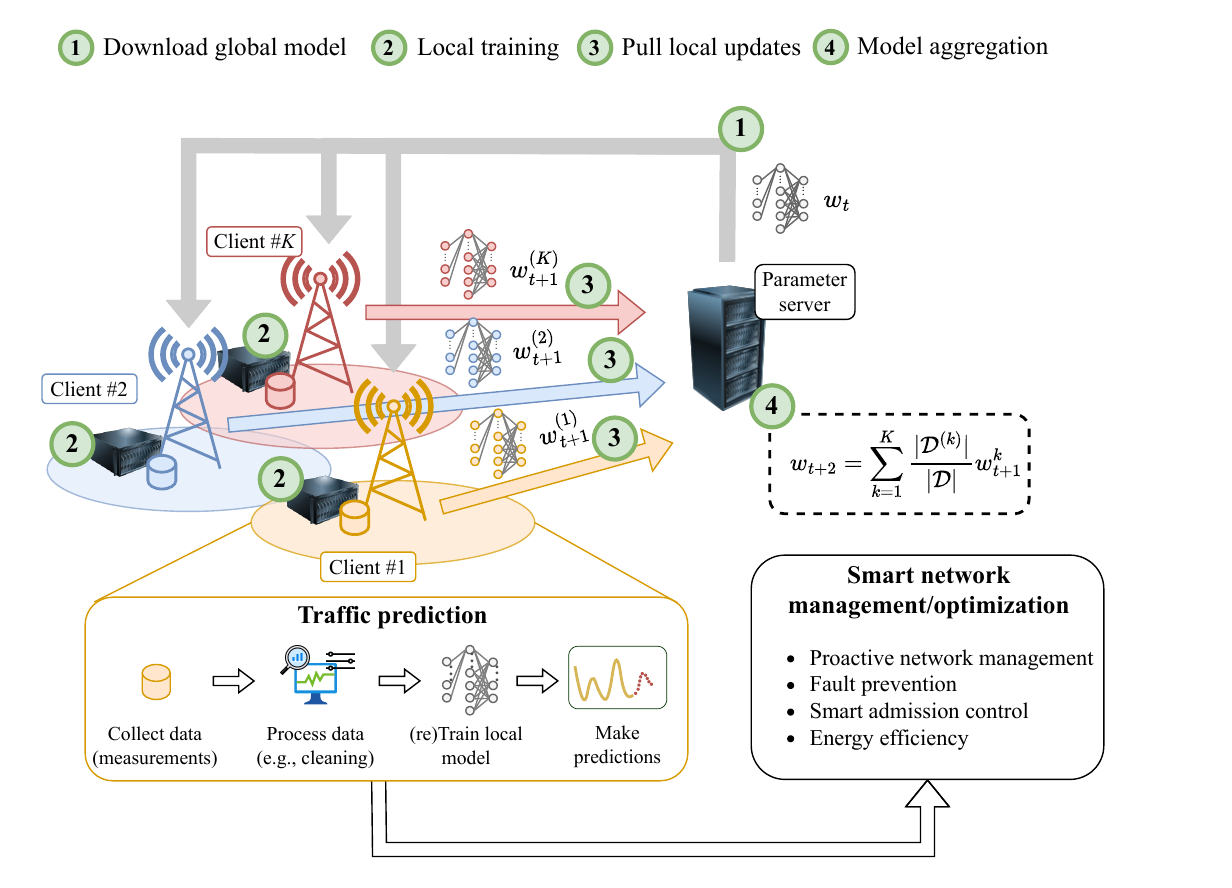}
    \caption{Overview of federated learning traffic prediction involving multiple BS sites. The steps corresponding to FL model training are marked with green circles.}
    \label{fig:fl_procedure}
\end{figure}

To derive the weights $w \in \mathbb{R}^d$ of the neural network model $f(\cdot)$ through FL, a set of clients $\mathcal{K}=\{1,2,...,K\}$ cooperate with local model updates $w^{(k)}, \forall k\in\mathcal{K}$. The goal is to
minimize a global loss function $F(w) = \frac{1}{K} \sum_{k=1}^{K} f^{(k)}(w)$, where $f^{(k)}(w)$ is the individual loss function of client $k$. A central MEC server, that is integrated in the cellular network, is in charge for orchestrating the operation of clients and performing model aggregations. In particular, through FL, an ML model is trained iteratively by performing the following steps (see Fig.~\ref{fig:fl_procedure}): 
\begin{enumerate}
    \item A subset of clients $\mathcal{K}' \subseteq \mathcal{K}$ is selected to participate in the current FL round, $t$. In the first steps, the selected clients download the global model $w_t$ from the parameter server. The method for selecting clients is further described in Section~\ref{subsec:fl_client_selection}.
    \item The subset of clients $\mathcal{K}'$ use the latest global model and their local datasets $\mathcal{D}^{(k)}$ to update the model weights as 
    \begin{equation}
        w_{t+1}^{(k)} \leftarrow w_{t}^{(k)} - \eta \nabla l^{(k)}(w_t,\mathcal{D}^{(k)}). 
    \end{equation}	
    \item The parameter server pulls the model updates computed by the selected subset of clients $\mathcal{K}'$.
    \item The server aggregates the received client models by following an aggregation strategy such as federated averaging (FedAvg)~\cite{mcmahan2017communication}. The model aggregation procedure is further described in Section~\ref{subsec:fl_client_selection}.
\end{enumerate} 

Model aggregation is a key component of FL, as it affects the convergence of the learning process and the overall predictive accuracy. Being $c^{(k)}$ the contribution of client $k$ (e.g., the proportional size of its local dataset compared to the entire distributed dataset), the global model is obtained as

\begin{equation}
    w = \frac{1}{\sum_{k=1}^{K} c^{(k)}} \sum_{k=1}^{K} c^{(k)} w^{(k)}.
\end{equation}

\subsection{Performance Evaluation Metrics}

\subsubsection{Prediction performance}
We use the mean absolute error ($\mathrm{MAE}$) and the normalized root mean squared error ($\mathrm{NRMSE}$) to measure the prediction error of the studied models. 

The MAE of a particular client $k$ is given by
\begin{equation}
    \mathrm{MAE}^{(k)} = \frac{1}{Td'} \sum_{t=1}^{T}\left(\sum_{i=1}^{d'} \lvert\hat{y}^{(k)}_{t,i} - y^{(k)}_{t,i} \rvert \right).
\end{equation}

Likewise, the $\mathrm{NRMSE}$ of client $k$ is defined as follows:

\begin{equation}
   \mathrm{NRMSE}^{(k)} = \frac{1}{\overline{y}^{(k)}}\sqrt{\frac{\sum_{t=1}^{T}\left(\sum_{i=1}^{d'} \left(\hat{y}^{(k)}_{t,i} - y^{(k)}_{t,i} \right)^2\right)}{Td'}},
\end{equation}
where $\overline{y}^{(k)}$ is the expectation over the true values $y^{(k)}$ of client $k$. Whenever we compare performance across various learning settings, we utilize the NRMSE due to its scale independence, facilitating a fair comparison. However, in reports concerning experiments exclusive to the FL setting, we opt for the MAE that effectively illustrates the quality predictions, showcasing the differences between the actual and the predicted values (in bits).

\subsubsection{Energy consumption model}

To better understand the accuracy-energy trade-off of the studied models, we use the sustainability indicator $S$ introduced in~\cite{perifanis2023towards}, which is applied to both centralized and federated settings. The lower the value of $S$, the better the trade-off between computational efficiency and predictive accuracy for the employed model. In particular,
\begin{equation}
    S = S_\text{Tr} \times S_\text{Inf},
    \label{eq:sustainability}
\end{equation}
where $S_\text{Tr}$ and $S_\text{Inf}$ represent the sustainability indicators calculated during the \textit{training} and \textit{inference} phases, respectively.

The sustainability of the training phase is as follows:
\begin{equation}\label{eq:strain}
    S_\text{Tr} = (1 + E_\text{Val})^{\alpha} \times (1 + C_\text{Tr})^{\beta} \times (1 + DS)^{\gamma},
\end{equation}
where $E_\text{Val}$ is the validation error, $C_\text{Tr}$ represents the total energy consumed for model training in Wh (we assume that the communication energy is negligible \cite{guerra2023cost}) and $DS$ is the data size to be transmitted to the central server in kB. Under the FL setting of our case study, $DS$ represents the model size that is transmitted per federated round. The exponents denote the importance of each value, with $\alpha + \beta + \gamma = 1$. 

As for the inference component, it is computed as
\begin{equation}\label{eq:sinference}
    S_\text{Inf} = E_\text{Test}^{\alpha'} \times C_\text{Inf}^{\beta'},
\end{equation}
where $E_\text{Test}$ is the error on the test data and $C_\text{Inf}$ is the energy consumed during inference. Similar to $S_\text{Tr}$, $\alpha' + \beta' = 1$.

The sustainability indicator $S$ was designed to compare different learning settings under controlled conditions, focusing on key factors that directly impact both the model's performance and resource usage. The indicator specifically incorporates:
\begin{enumerate}
    \item Accuracy on unseen data, measured by the prediction error.
    \item Computational efficiency with respect to the energy consumed in Watt hours (Wh).
    \item Communication efficiency, quantified by the data size to be transmitted in kilobytes (kB).
\end{enumerate}
These components were selected because they provide a comprehensive view of the trade-offs between model performance and resource efficiency. Other aspects such as training epochs and federated rounds were kept consistent across different experiments to facilitate a fair comparison.

\subsection{Proposed LSTM-based Solution}
In this section, we describe our LSTM based multi-output and multi-step time series forecasting network $f(\cdot)$. The selection of the LSTM model is based on the comprehensive analysis conducted by Perfanis et al. \cite{perifanis2023federated}, which highlighted the model's high predictive accuracy and robustness in similar contexts against other well-known models such as RNN, CNN, GRU, or Transformer. LSTM models offer a more balanced approach, providing sufficient accuracy while minimizing communication overhead, which is crucial for maintaining energy efficiency in sustainable telecommunication networks.

At $k^\mathrm{th}$ base-station, our aim is to predict $y_t^{(k)} \in \real^{d'}$ using previous samples and the prediction network, i.e., $\hat{y}_T^{(k)} = f( x_{1}^{(k)},\cdots,x_{t-1}^{(k)} )$. We consider deep networks where $M$ LSTM layers, followed by a feed-forward neural network with $L$ layers. We define $m^\mathrm{th}$ layer of the LSTM that uses the formulation in~\cite{hochreiter1997long} as:
\begin{align*}
\vec{z}_t\m &= \mathrm{tanh}(\vec{W}\m_z \h_t^{(m-1)} + \vec{V}\m_z \vec{h}\m_{t-1} + \vec{b}_z\m) \\
\vec{s}_t\m &= \mathrm{sigmoid}(\vec{W}\m_s \h_t^{(m-1)} + \vec{V}\m_s \vec{h}\m_{t-1} + \vec{b}_s\m) \\
\vec{f}_t\m &= \mathrm{sigmoid}(\vec{W}\m_f \h_t^{(m-1)} + \vec{V}\m_f \vec{h}\m_{t-1} + \vec{b}\m_f) \\
\vec{c}_t\m &= \vec{s}_t\m \odot \vec{z}_t\m + \vec{f}_t\m \odot \vec{c}_{t-1}\m \\
\vec{o}_t\m &= \mathrm{sigmoid}(\vec{W}\m_o\h_t^{(m-1)}+\vec{R}\m_o \vec{h}_{t-1}\m+ \vec{b}\m_o) \\
\vec{h}_{t}\m &= \vec{o}_t\m \odot \mathrm{tanh}(\vec{c}_t\m), 
\end{align*}
where $\h_t^{(0)}=x_t$, $\h_0\m \sim \mathcal{N}(0, 0.01)$, $\vec{c}_t\m \in \real^m$ is the cell state vector, $\vec{h}_t\m \in \real^w$ is the hidden state vector, for the $t$\textsuperscript{th} LSTM unit. $\vec{s}_t\m$, $\vec{f}_t\m$ and $\vec{o}_t\m$ are the input, forget and output gates, respectively. $\odot$ is the operation for elementwise multiplication. $\vec{W}$, $\vec{V}$, and $\vec{b}$ with the subscripts $z$, $s$, $f$, and $o$ are the parameters of the LSTM unit to be learned. 

We then input final layer LSTM output $\vec{h}_t^{(M)}$ to a feed forward neural network of $L$ layers, where $l^\mathrm{th}$ layer with weight $\vec{W}_l$ multiplies the preceeding layer's output, followed by $\mathrm{ReLU}$ activation function, as:
\begin{equation}
    \vec{v}^{(l)} = \mathrm{ReLU}(\vec{W}_{l-1} \vec{v}^{(l-1)}),
\end{equation}
where $\vec{v}^{(0)} = \vec{h}_T^{(M)}$.

Lastly, we project intermediate representation into predictions using a linear layer with weight $\vec{W}_p$ as:
\begin{equation}
    \hat{y}_t^{(k)} = \vec{W}_p \vec{v}^{(L)},
\end{equation}
where $\hat{y}_t^{(k)} \in \real^{d'}$.

\section{Dataset Description and Analysis}
\label{section:dataset}

\subsection{Dataset Description}

The dataset used for this case study was gathered from five different locations in Barcelona (Spain), thus aiming to provide unique network utilization patterns, including daily living and especial events. The set of studied locations are the following:
\begin{itemize}
    \item \textbf{Les Corts - Camp Nou (LCCN):} A residential area nearby Camp Nou (\textit{Football Club Barcelona}'s stadium), which regularly hosts soccer matches and other special events. Measurements at this location comprise 12 days (2019-01-12 to 2019-01-24), including three soccer matches.
    \item \textbf{Poble Sec (PS):} A residential and touristic area enclosed between strategic points, including the historic center, the mountain of \textit{Montjuïc}, the exhibition centre and the port. Measurements at this location comprise 28 days (2018-02-05 to 2018-03-05).
    \item \textbf{El Born (EB):} A touristic area in the downtown of the city. It is characterized by having a lot of amusement and nightlife. Measurements at this location comprise 7 days (2018-03-28 to 2018-04-04).
    \item \textbf{Sants (S):} A residential area with the biggest train station of the city. Measurements were collected at two distinct temporal intervals, thus leading to splits S1 and S2, jointly spanning a duration of 58 days (2021-08-03 to 2021-09-09 and 2021-09-09 to 2021-09-29).
    \item \textbf{Eixample (E):} Located in the heart of the city, this is a residential area with relevant touristic interest points (e.g., \textit{La Sagrada Familia}). Data from this location was acquired during two separate time periods, thus leading to splits E1 and E2, with a total duration of 58 days (2021-11-17 to 2021-11-29 and 2021-12-11 to 2022-01-15).
\end{itemize}

The measurements were retrieved from downlink control information (DCI) messages, which are transmitted through the PDCCH at every transmission time interval (TTI), e.g., every one millisecond. The resulting features are described in Table~\ref{tab:features_dataset}, which includes complementary features like the uplink/downlink throughput. The dataset has been partitioned into training ($D_{tr}$), validation ($D_{val}$), and test ($D_{test}$) sets. Erroneous values due to decodification errors were eliminated (by zeroing corrupted samples) from the dataset as part of the general cleansing performed. Moreover, we downsampled the data by averaging non-overlapping two-minute intervals (i.e., 120 consecutive samples), aligning with our goal of predicting long-term effects rather than capturing high-precision fluctuations. Finally, a normalization step (standard scaler) has been applied to eliminate the influence of value ranges.

\begin{table}[ht!]
\centering
\caption{Set of features captured in the LTE PDCCH dataset.}
\label{tab:features_dataset}
\resizebox{\columnwidth}{!}{\begin{tabular}{@{}cl@{}}
\toprule
\textbf{\textbf{Feature}} & \multicolumn{1}{c}{\textbf{Description}} \\ \midrule
$\bar{\text{RB}}_{dl}$ & The average number of allocated resource blocks in the DL\\
$\sigma^2({\text{RB}}_{dl})$ & The normalized variance of $\text{RB}_{dl}$. \\
$\bar{\text{RB}}_{ul}$ & The average number of allocated resource blocks in the UL \\
$\sigma^2({\text{RB}}_{ul})$ & The normalized variance of $\text{RB}_{ul}$. \\
$\text{RNTI}_c$ & \begin{tabular}[c]{@{}l@{}}The average RNTI counter, indicating the average number\\ of users observed, during the selected time window.\end{tabular} \\
$\bar{\text{MCS}}_{dl}$ & \begin{tabular}[c]{@{}l@{}}The average Modulation and Coding Scheme (MCS)\\ index in the downlink (in 0-31).\end{tabular} \\
$\sigma^2({\text{MCS}}_{dl})$ & The normalized variance of the MCS index in the downlink. \\
$\bar{\text{MCS}}_{ul}$ & The average MCS index in the uplink (in 0-31). \\
$\sigma^2({\text{MCS}}_{ul})$ & The normalized variance of the MCS index in the uplink. \\
$\text{TB}_{dl}$ & The downlink transport block size in bits, according to~\cite{TS36213}. \\
$\text{TB}_{ul}$ & The uplink transport block size in bits, according to~\cite{TS36213}. \\ \bottomrule
\end{tabular}%
}
\end{table}

\subsection{Dataset Analysis}

Our analysis begins with a detailed examination of the data distributions in each BS. For that, we measure the differences in feature distributions among the BSs. For that, we first convert each feature from the time series of each BS into a histogram. Then, we compute the Kullback-Leibler (KL) divergence to quantify the similarity in feature distributions between pairs of BSs:
\begin{equation}
    D_{KL}(P \parallel Q) = \sum_{x \in X} P(x) \log\left(\frac{P(x)}{Q(x)}\right).
\end{equation}

\begin{figure}[ht]
    \centering  
    \includegraphics[width=.75\columnwidth]{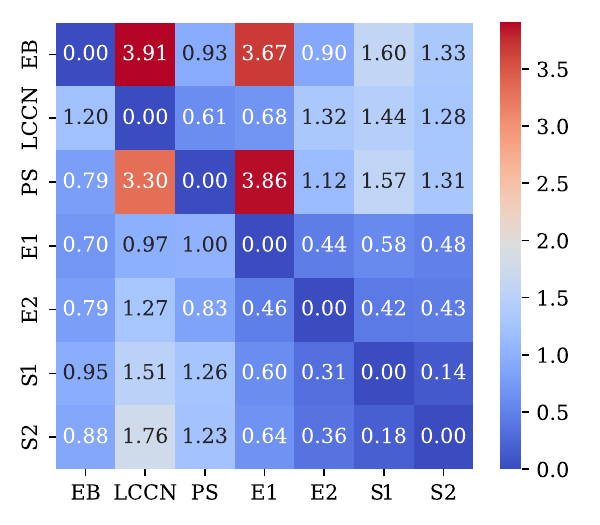}
    \caption{Average KL-divergence between datasets from different BSs.}
    \label{fig:kl_map}
\end{figure}

Figure \ref{fig:kl_map} depicts the average KL divergence between pairs of BSs, considering all features. Note that KL is not symmetric, i.e., $D_{KL}(P \parallel Q) \neq D_{KL}(Q \parallel P)$. We focus on pairings exhibiting high KL values, specifically between EB-LCCN, EB-E1, PS-LCCN, and PS-E1. These high values are relevant to the FL operation, as they could indicate substantial disagreement among clients. In the sequel, we further evidence how high KL values lead to high discrepancies in terms of predictive accuracy, emphasizing the necessity to address non independent and identically distributed (non-iid) data distributions within federated settings.

\begin{figure}[ht]
    \centering  \includegraphics[width=\columnwidth]{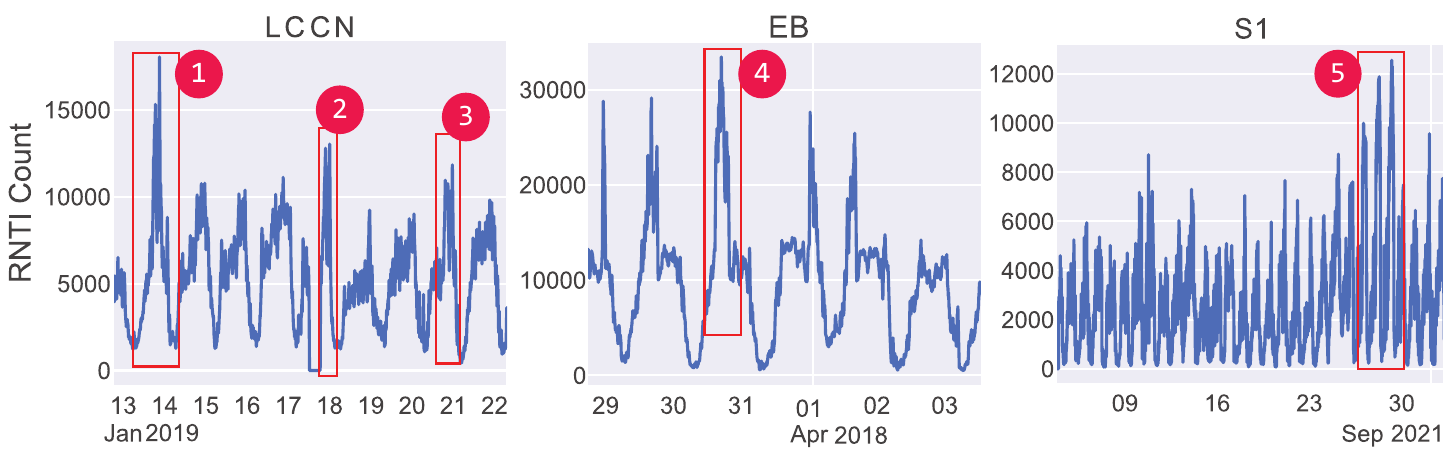}
    \caption{Special events that affect the RNTI count in LCCN, EB, and S1.}
    \label{fig:special_events}
\end{figure}

Differences in data distribution across BSs can be linked to specific events within the spatio-temporal context of the data collection process. To provide further insights into the nature of the data at the different BSs, we have identified real-world incidents that align with the data collection periods near the monitoring BSs. Recognizing these distinct patterns/anomalies, associated with real-world occurrences, highlights the effectiveness of our data collection methodology. Figure~\ref{fig:special_events} illustrates the fluctuations in the RNTI count variable due to these events at three base stations. Starting with the left-most plot (at LCCN), there are three notorious peaks in the network utilization, which correspond to three football matches: (1) FC Barcelona vs SD Eibar (2019-01-13), with 71,039 spectators, (2) FC Barcelona vs Levante UD (2019-01-17), with 42,838 spectators, and (3) FC Barcelona vs CD Leganés (2019-01-2), with 50,670 spectators. When it comes to EB (middle plot), the Good Friday led to possible crowds in the region due to the Barcelona Cathedral service (4). Finally, the right-most plot shows that the BS in S1 was heavily used between 2021-08-24 and 2021-08-29 (5), which matches with the local festivities (Festa Major de Sants) that attract many visitors due to different types of activities.

\section{Performance Evaluation}
\label{section:performance_evaluation}

In this section, we study the application of FL to traffic prediction by evaluating a set of models and techniques using real LTE data measurements. The experiments aim to provide insights on different aspects of the downstream ML task and at the same time explore the advantages of FL over centralized solutions, including both predictive performance and environmental impact. To such end, we ran the complete set of experiments in a local environment composed of a Windows~11 workstation equipped with an AMD Ryzen 5 4600H CPU and 16 GB memory. This setup resembles real-world scenarios where BS edge servers have moderate computation capabilities and cannot devote all of their power to ML tasks since they already need to run the network. To ensure the validity of the results, 10 different seeds were used for the initialization of random generators.\footnote{The source code used in this paper is available at \url{https://github.com/vperifan/Federated-Time-Series-Forecasting}, acc. on Feb. 25 2024.} The ML models were implemented in PyTorch and the measurements on energy consumption and CO$_2$ emissions were obtained using CarbonTracker~\cite{anthony2020carbontracker}, which is a library that allows measuring the power consumed by GPU, CPU, and DRAM devices when running ML model training and inference. Table~\ref{tab:eval_param} gathers the evaluation parameters and the selected hyperparameters for model training.

\begin{table}[ht!]
\centering
\caption{Evaluation parameters.}
\label{tab:eval_param}
\resizebox{\columnwidth}{!}{%
\begin{tabular}{@{}clc@{}}
\toprule
\textbf{Parameter} & \multicolumn{1}{c}{\textbf{Description}} & \textbf{Value} \\ \midrule
$Seeds$ & Number of seeds per experiment & 10 \\
$R$ & Number of FL rounds & 10 \\
$E$ & Number of epochs per FL round & 3 \\
$T$ & Prediction step & [1-10] \\
$p(D_{tr})$ & Percentage of data used for training & 60\% \\
$p(D_{val})$ & Percentage of data used for validation & 20\% \\
$p(D_{test})$ & Percentage of data used for test & 20\% \\
$f(\cdot)$ & Neural Network model & LSTM \\
$d$ & Input feature dimension & 11 \\
$d'$ & Output tensor (predictions) dimension & 5 \\
$[\alpha, \beta, \gamma]$ & Sustainability Indicator Weights (Train) & [0.33, 0.33, 0.33] \\
$[\alpha', \beta']$ & Sustainability Indicator Weights (Inf) & [0.5, 0.5] \\ \bottomrule
\end{tabular}%
}
\end{table}

\subsection{Cooperate, Not to Cooperate, or Centralize?}

\begin{table*}[t!]
\caption{Average and standard deviation of the test error (NRMSE) achieved by the LSTM model per BS and in average, for each considered model training approach (individual, centralized, and federated). The total energy consumption is also included.}
\label{tab:settings_overview}
\begin{adjustbox}{width=\textwidth,center}
\begin{tabular}{c|c|c|c|c|c|c|c|c|c}
 &
  EB &
  LCCN &
  PS &
  E1 &
  E2 &
  S1 &
  S2 &
  Avg. &
  \begin{tabular}[c]{@{}c@{}}Total Energy\\ Consumption (Wh)\end{tabular} \\ \hline
Individual  & 2.442 ± 0.014 & 2.764 ± 0.01 & 1.112 ± 0.025 & 2.108 ± 0.091  & 1.02 ± 0.013 & 2.924 ± 0.003 & 1.061 ± 0.004 & 1.92 ± 0.0235  & 13 \\ \hline
Centralized & 1.06 ± 0.008 & 0.725 ± 0.006 & 1.092 ± 0.006 & 2.295 ± 0.022 & 1.39 ± 0.047 & 2.490 ± 0.002 & 0.985 ± 0.009 & 1.434 ± 0.003 & 15.5 \\ \hline
Federated &
  0.54 ± 0.004 &
  0.975 ± 0.03 &
  1.468 ± 0.001 &
  2.027 ± 0.014 &
  1.062 ± 0.007 &
  2.625 ± 0.015 &
  1.001 ± 0.028 &
  \textbf{1.385} ± 0.044 &
  14 \\ \hline
\end{tabular}
\end{adjustbox}
\end{table*}

We start our analysis by comparing the predictive performance under three different learning settings: \textit{Individual}, \textit{Centralized}, and \textit{Federated}. The \textit{Individual} setting entails that each BS trains an independent model using only its locally available data. In contrast, in the \textit{Centralized} setting, a central server uses the data from all the subscribed BSs to train a single model. Ultimately, the \textit{Federated} setting allows the BSs to train a single collaborative model without exchanging their local datasets. The experiments were conducted based on an equal number of dataset accesses by the algorithm to ensure a fair and balanced comparison among different settings. While the number of accesses is equal to the number of training epochs in the individual and centralized settings, in the federated one, it is obtained as the product of the local training epochs by the number of federated rounds ($R \times E$).
Results from our experiments, are presented in Table~\ref{tab:settings_overview}, which shows the average and standard deviation of the test accuracy achieved by the different approaches in each BS. Apart from that, the average performance and the total energy consumption (Wh for training) are provided.

In the \textit{individual} setting, the average error shows a significant variance across BSs, with the lowest and highest values observed in E2 (1.02) and S1 (2.924), respectively. This setting reflects moderate prediction accuracy with considerable fluctuations in the performance across different BSs. Notably, under this setting, the lowest energy consumption at 13.032 Wh is recorded, mainly due to the lack of communication among BS. Under the \textit{centralized} setting, some BSs exhibit improved prediction accuracy (e.g., EB and LCCN) while others perform worse than in the individual setting, lacking generalization ability. Yet, the centralized provides a better average performance than the individual, indicating that pooling resources and data at a central point enhances overall prediction accuracy. However, this setting leads to the highest energy consumption (19\% higher than the individual setting), which can be attributed to the centralized processing demands of larger amount of data, indicating that, while prediction accuracy is enhanced, it comes at the cost of increased energy consumption. Finally, the \textit{federated} setting stands out by combining the advantages of both aforementioned settings. It achieves the lowest average error of 1.385 through mixed but generally improved results across BSs. The energy consumption in this setting is 14 Wh, which is around 8\% higher than the individual setting but 10\% lower than the centralized one.

\begin{figure}[ht!]
    \centering        \includegraphics[width=\columnwidth]{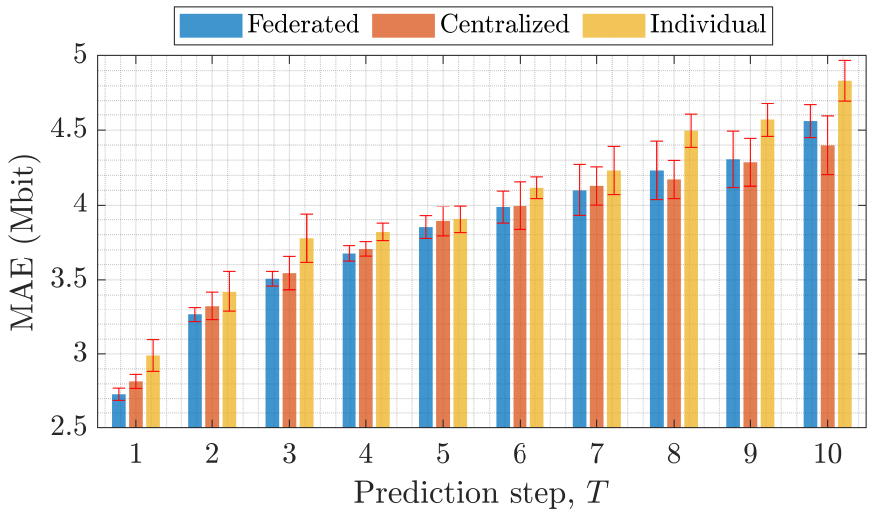}
    \caption{Mean and standard deviation of the MAE obtained by each learning approach, for different prediction step values $T\in[1-10]$.}
    \label{fig:prediction_steps}
\end{figure}

To further analyze the predictive performance of the different approaches, we show the performance achieved at an increasing number of future time steps in Fig.~\ref{fig:prediction_steps}, which includes the error at $T\in [1,10]$. As shown, the general trend is that the error increases as the prediction targets a more distant point in time. In particular, as the forecast horizon expands, the influence of immediate past traffic conditions on future states diminishes, making the prediction task more susceptible to unforeseen fluctuations in network usage and external factors affecting user behavior. It is worth highlighting the differences between learning settings. As it can be observed, centralized learning demonstrates an interesting pattern. In the beginning, it starts with a higher MAE compared to FL. However, as the prediction steps increase the centralized MAE increases at a reduced rate, until step 8 when it achieves a lower error than FL. This is possibly due to the larger amount of data that is available for training in the centralized learning, which enables the model to capture long-range dependencies in time. Individual learning has the worst performance compared both to FL and centralized learning for each considered time step. Thus, it is clear that it is crucial to optimize the trade-off between future prediction steps and acceptable levels of predictive accuracy under specific learning settings, aiming to successfully perform resource allocation and planning.

Next, to reinforce the suitability of FL, Table~\ref{tab:sustainability} provides the sustainability performance of all the settings based on $S$ (see Eq.~\eqref{eq:sustainability}), where FL's superiority over other settings is proven. Specifically, FL gets the lowest $S$ value ($S=30.24$), thus achieving the best trade-off between predictive accuracy and computational and communication efficiency. Individual learning comes second with $S=71.67$ over $S=132.03$ in the centralized learning.

In our experiments, federated learning slightly outperforms centralized learning, which can be attributed to the unique characteristics of network traffic data. The decentralized setting allows local models to capture site-specific patterns and nuances in traffic behavior, which may not be fully captured in a centralized approach where these variations are averaged out. By training on localized data, federated learning can better adapt to the diverse patterns observed at different base stations, particularly in non-IID environments, where the combination of human-driven and machine-driven processes adds complexity. Unlike image domain datasets, timeseries forecasting demonstrates unique behavior in decentralized settings, as temporal dependencies and varying trends across clients introduce distinct challenges and opportunities for federated models to capture local temporal patterns more effectively. This can be significant beneficial, especially if considering the extra sustainability benefits that federated learning brings.
It is clear that although centralized learning provides a good level of accuracy, it comes at the cost of excessive energy consumption during training, as well as the extra costs associated with the high volume of data that needs to be transmitted. Conversely, individual learning may have low costs regarding energy consumption and data transmission; however, it demonstrates a higher prediction error, probably due to an inability to generalize. Therefore, the federated approach effectively balances the trade-off between computational efficiency and prediction accuracy by distributing the learning process across the participating BSs. Despite a minor increase in energy consumption compared to the individual setting, the gains in prediction accuracy justify this approach, making the federated setting the most effective and balanced choice for the downstream task.

\begin{table}[ht!]
\caption{Sustainability performance ($S$) achieved by the LSTM model under the different considered model training approaches.}
\label{tab:sustainability}
\begin{adjustbox}{width=\columnwidth,center}
\begin{tabular}{@{}cccccccc@{}}
\toprule
 & NRMSE & $E_{Tr}$ (Wh) & $E_{Inf}$ (Wh) & Size (kB) & $S_{Tr}$ & $S_{Inf}$ & $S$ \\ \midrule
Ind. & 1.92 & 13.03 & 0.048 & NA & 40.98 & 1.75 & 71.68 \\
Cen. & 1.43 & 15.5 & 0.03 & 16531 & 83.43 & 1.58 & 132.03 \\
Fed. & 1.38 & 14.06 & 0.03 & 217 & 19.27 & 1.57 & 30.24 \\ \bottomrule
\end{tabular}%
\end{adjustbox}
\end{table}

\subsection{Federated Learning Fine-Tuning}
We next investigate the further fine-tuning of the FL setup for enhancing its overall predictive performance. For that, we focus on outliers handling and model aggregation. 

\subsubsection{Outliers Handling}

\begin{figure}[ht!]
    \centering    
    \includegraphics[width=\columnwidth]{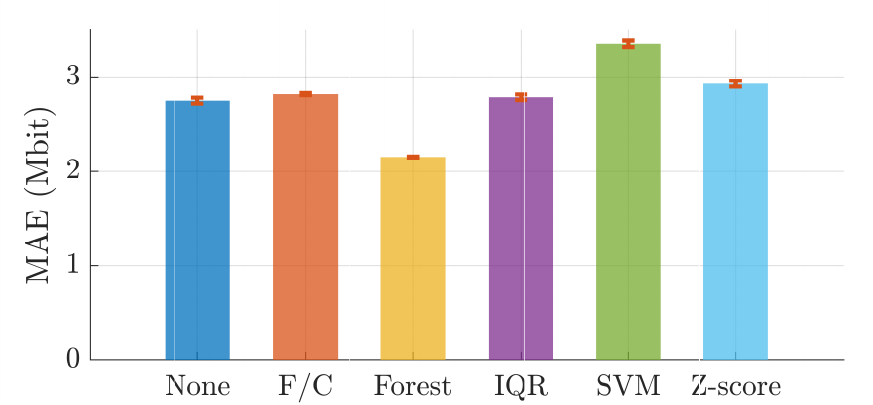}
    \caption{MAE achieved by applying each considered outlier detection and correction method.}
    \label{fig:outliers}
\end{figure}

A key aspect of data preprocessing is outlier detection and their removal to improve the effectiveness of the model training. To comprehensively compare different outlier detection approaches, in Fig.~\ref{fig:outliers}, we employ during LSTM training and compare flooring/capping (F/C), isolation forest (Forest)~\cite{liu2008isolation}, interquartile range (IQR)~\cite{rousseeuw1993iqr}, support vector machine (SVM)~\cite{jordaan2004svm} and Z-score. In the context of our study which focuses on time series forecasting, it is not feasible to directly remove outliers from the dataset. This is  because an elimation will break the integrity of the temporal sequence, i.e., gaps will be created, which subsequently mislead the analysis and the forecasting model's understanding of historical patterns. To preserve the sequence, following outlier detection, we employ the flooring and capping technique. Specifically, we apply flooring to outliers below the minimum threshold and capping to outliers above the maximum threshold. 
It is important to note that this outlier correction is applied only to the training dataset. The reason for this is to prevent the model from being influenced by extreme values during training while still allowing the model to evaluate on the true test data distribution, including any potential outliers.
The results reveal that the isolation forest technique outperforms the rest of the approaches in terms of improving forecasting accuracy. Contrary to other outlier detection methods that focus on profiling normal data points, isolation forest leverages the assumption that outliers are inherently few and exhibit distinct characteristics from the majority. This approach facilitates a more effective identification of anomalies, particularly in cellular traffic data, where outliers result from unforeseen events or data collection inaccuracies can significantly deviate from typical traffic patterns. 
By correctly identifying and handling these anomalies during training, isolation forest helps the model better understand the underlying patterns in the data, leading to improved forecasting accuracy.
Interestingly, methods such as flooring/capping, IQR, SVM, and Z-score perform worse than the baseline, where no outlier detection and mitigation are applied. This underperformance is primarily due to the nature of time series data. Traditional outlier detection methods~\cite{blazquez2021review} are designed for static datasets and often struggle with time series because they do not consider the temporal dependency between data points. These methods may either fail to identify context-dependent anomalies or incorrectly flag normal variations in the time series as outliers. This can distort the data distribution, leading to models that are poorly calibrated and ultimately less accurate.

\subsubsection{Model Aggregation}

Model aggregation is a key component of every FL system. \textit{Federated averaging (FedAvg)}~\cite{mcmahan2017communication} is the most widely used aggregator in FL applications due to its simplicity and robustness. Despite being simple enough and performing well in general, FedAvg~\cite{mcmahan2017communication} might cause objective inconsistency in certain scenarios. This can happen due to non-iid and heterogeneous data. To mitigate this risk, alternative methods have emerged to better accommodate data diversity. Wang et al.~\cite{wang2020fednova} introduced FedNova, which adapts the aggregation process by normalizing local model updates. This adjustment involves dividing local gradients by the number of client steps before their aggregation, rather than just directly averaging their local gradients. FedAvgM, proposed by Tzu et al.~\cite{tzu2020fedavgm}, enhances FedAvg by incorporating server momentum, which involves applying a hyper-parameter $\beta$ to previous model updates during each training epoch before adding new updates, thereby refining the aggregation process. In addition, there have been developments in federated versions of adaptive optimizers, such as FedAdagrad, FedYogi, and FedAdam, as discussed in \cite{reddi2020adaptive}. These techniques are designed to handle heterogeneous data, enhance model performance, and reduce communication overhead.

\begin{figure}[ht!]
    \centering   
    \includegraphics[width=.9\columnwidth]{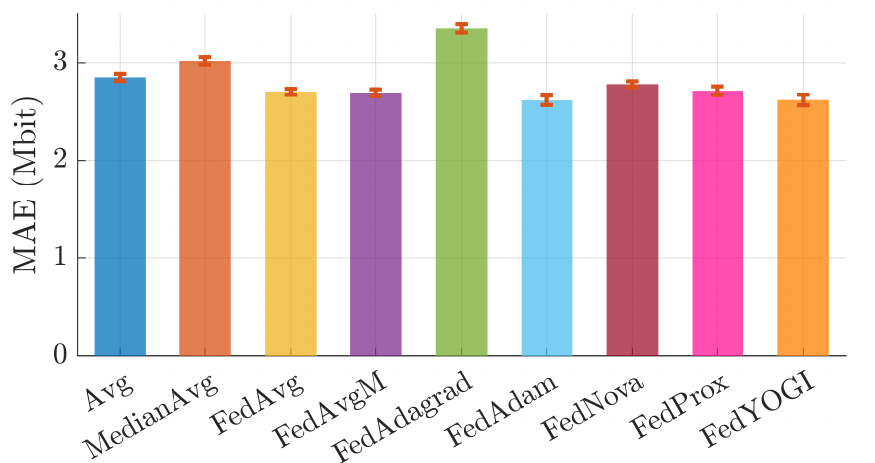}
    \caption{Mean and standard deviation of the MAE achieved by each model aggregation strategy.}
    \label{fig:aggregators}
\end{figure}

In our experiments, we evaluated nine aggregation functions to address the challenges posed by quantity, quality, and temporal skew in our dataset. Our analysis, illustrated in Fig.~\ref{fig:aggregators}, underscores the superior performance of FedAdam over its counterparts, but with FedAvg coming second, offering similar performance and with lower deviation across experiments. This outcome suggests that, while specialized algorithms like FedAdam are designed to mitigate the effects of non-iid data distributions through adaptive normalization, their optimal performance hinges on precise hyperparameter tuning tailored to the specific dataset characteristics. Conversely, the relatively simple FedAvg algorithm, despite its straightforward approach, demonstrates robustness and effectiveness, indicating its potential applicability across a broader range of FL scenarios without extensive customization. Since we did not observe significant differences, we utilized the FedAvg algorithm for model aggregation for generality and simplicity.

\subsection{Impact of Individual Contributions in Client Selection}
\label{subsec:fl_client_selection}

We now provide insights on the potential of FL client selection by analyzing the contributions of each client to the predictive performance of the federated model. For that, Table~\ref{tab:selection_brute_force} shows the test error obtained at the different BSs for various settings where different subsets of BSs are available to be selected by the FL central server (ranging from 1 to 7 clients selected randomly per round). The results are compared to the cases where a single BS is available to be selected, which resembles to the individual training setting studied above.

The results reveal that the selection and number of clients participating in each FL round leads to a significant variance in the global model's prediction error on the test sets of the different clients. In general, it can be observed that, as the number of clients participating in each round decreases (from 7/7 to 1/7), the error across all the BSs increases, indicating that having more clients participating in each round tends to improve prediction accuracy. However, it is worth highlighting that the ML model has an adequate performance even with 6/7 or 5/7 participating clients, with only a small degradation of 4-5\% compared to the scenario that considers all the BSs for training. These results advocate the use of a client selection mechanism that could lead to an adequate level of performance by utilizing fewer clients.


To better understand the contribution of each BS to the global model, we conducted additional experiments using a deletion-based scheme, where one BS was excluded from the FL training process at a time. The results are also presented in Table \ref{tab:selection_brute_force}, indicate that while the overall average error may not be significantly impacted, the exclusion of certain BSs can lead to notable performance degradation on specific test sets. For example, the exclusion of E2 resulted in a greater increase in error on S1 compared to the exclusion of S1 itself. This phenomenon can be explained by the low KL-Divergence between the datasets of E2 and S1, as illustrated in Figure \ref{fig:kl_map}, suggesting a close similarity in data distribution between these clients.

These observations underscore the potential benefits of a strategic client selection mechanism that favors clients with contributions that significantly enhance the global model's generalization ability. This can be particularly beneficial if our aim is to minimize energy costs associated with the training of ML models, especially in a Traffic Forecasting scenario that might scale fast in a real-world application to incorporate thousands of BS.

However, the endeavor to optimize client selection for FL not only aims to bolster predictive performance but also seeks to expedite convergence and minimize the environmental footprint of the training process. The challenge thus extends beyond recognizing the contribution of individual clients to the model's accuracy, encompassing the need to efficiently orchestrate the training process. FL client selection mechanisms targeting high-performing nodes are key to boosting overall performance.

\begin{table}[ht!]
\caption{Test error (NRMSE) experienced by each BS for the different combinations of client selections.}
\begin{adjustbox}{width=\columnwidth,center}
\begin{tabular}{@{}cccccccccc@{}}
\toprule
\multirow{2}{*}{\textbf{BS avail.}} &
  \multirow{2}{*}{\textbf{\begin{tabular}[c]{@{}c@{}}\# Clients \\ per round\end{tabular}}} &
  \multicolumn{8}{c}{\textbf{Global model's test error}} \\ \cline{3-10} 
               &                          & EB   & LCCN & PS   & E1   & E2   & S1   & \multicolumn{1}{c|}{S2}   & Average       \\ \hline
\multirow{7}{*}{All (7)} &
  \multicolumn{1}{c|}{7/7} &
  0.54 &
  0.98 &
  1.47 &
  2.03 &
  1.06 &
  2.62 &
  \multicolumn{1}{c|}{1.00} &
  \textbf{1.39} \\
               & \multicolumn{1}{c|}{6/7} & 0.48 & 0.86 & 1.48 & 1.90 & 1.20 & 3.12 & \multicolumn{1}{c|}{1.09} & 1.45          \\
               & \multicolumn{1}{c|}{5/7} & 0.46 & 0.87 & 1.42 & 1.91 & 1.23 & 3.21 & \multicolumn{1}{c|}{1.11} & 1.46          \\
               & \multicolumn{1}{c|}{4/7} & 0.54 & 0.99 & 1.35 & 1.95 & 1.11 & 2.97 & \multicolumn{1}{c|}{1.01} & 1.42          \\
               & \multicolumn{1}{c|}{3/7} & 0.72 & 1.20 & 1.53 & 2.04 & 1.16 & 2.69 & \multicolumn{1}{c|}{1.03} & 1.48          \\
               & \multicolumn{1}{c|}{2/7} & 1.00 & 1.09 & 1.50 & 1.96 & 1.71 & 3.30 & \multicolumn{1}{c|}{1.11} & 1.67          \\
               & \multicolumn{1}{c|}{1/7} & 1.09 & 0.99 & 1.66 & 2.07 & 1.92 & 3.16 & \multicolumn{1}{c|}{1.17} & 1.72          \\ 
               

\hline
All$\setminus$\{EB\}   & \multicolumn{1}{c|}{6/6} & 1.38 & 0.96 & 1.47 & 2.04 & 1.05 & 2.63 & \multicolumn{1}{c|}{1.00} & 1.50          \\
All$\setminus$\{LCCN\} & \multicolumn{1}{c|}{6/6} & 0.53 & 1.59 & 1.41 & 2.04 & 1.08 & 2.72 & \multicolumn{1}{c|}{1.01} & 1.48          \\
All$\setminus$\{PS\}   & \multicolumn{1}{c|}{6/6} & 0.84 & 0.83 & 2.29 & 2.02 & 1.09 & 2.60 & \multicolumn{1}{c|}{1.00} & 1.52          \\
All$\setminus$\{E1\}   & \multicolumn{1}{c|}{6/6} & 0.55 & 0.98 & 1.44 & 2.77 & 1.04 & 2.68 & \multicolumn{1}{c|}{0.99} & 1.49          \\
All$\setminus$\{E2\}   & \multicolumn{1}{c|}{6/6} & 0.45 & 0.82 & 1.48 & 1.92 & 2.48 & 3.11 & \multicolumn{1}{c|}{1.08} & 1.62          \\
All$\setminus$\{S1\} &
  \multicolumn{1}{c|}{6/6} &
  0.57 &
  0.85 &
  1.37 &
  1.93 &
  1.31 &
  2.81 &
  \multicolumn{1}{c|}{0.98} &
  \textbf{1.40} \\
All$\setminus$\{S2\}   & \multicolumn{1}{c|}{6/6} & 0.59 & 1.11 & 1.49 & 2.08 & 1.07 & 2.67 & \multicolumn{1}{c|}{2.13} & 1.59          \\  \bottomrule
\end{tabular}%
\end{adjustbox}
\label{tab:selection_brute_force}
\end{table}

\subsection{Personalized Federated Learning}

Aiming to investigate personalization aspects in FL~\cite{jiang2023personalisation}, we employ the technique of local fine-tuning (LF) to enhance the accuracy of the local models on the respective validation sets. LF involves an additional step to federated model training, where each participating client performs a complementary training round to fit the global model on their local data before deploying it for inference. The rationale behind LF is to allow the global model, which results from the aggregation of diverse local models, to better adapt to the specific characteristics and distribution of each client's local dataset. In the studied use case, the customization provided by LF ensures that the models used are better attuned to the patterns unique to each BS.

The results presented in Table~\ref{tab:lf-impact} illustrate the efficacy of LF in reducing MAE across different BSs, for both centralized and federated settings. As shown, in the centralized and federated settings, LF achieves 7.8\% and 10.9\%  average improvements, respectively. Such gains reveal the ability of LF to leverage the unique data characteristics of each client, which is a particularly compelling property when dealing with complex non-iid data. In conclusion, LF ensures that the model is not only generalized to perform well across the entire network but also optimized for specific local conditions.

\begin{table}[ht!]
\caption{Improvements of LF in the test error (NRMSE) in each BS, for each the centralized and federated approaches.}
\centering
\resizebox{\columnwidth}{!}{%
\begin{tabular}{@{}ccccccccc@{}}
\toprule
& EB & LCCN & PS & E1 & E2 & S1 & S2 & Avg. \\ \midrule
Cen-LF & 15.52\% & 18\% & 19.33\% & 1.12\% & 3.01\% & 5.73\% & 1.86\% & 7.84\% \\
Fed-LF & 31.36\% & 2.1\% & 1.01\% & 9.95\% & 16.5\% & 9.12\% & 17.06\% & 10.9\% \\ \bottomrule
\end{tabular}%
}
\label{tab:lf-impact}
\end{table}

\subsection{Combining Network and Exogenous Data Sources}

\begin{figure}[ht!]
    \centering      \includegraphics[width=0.8\columnwidth]{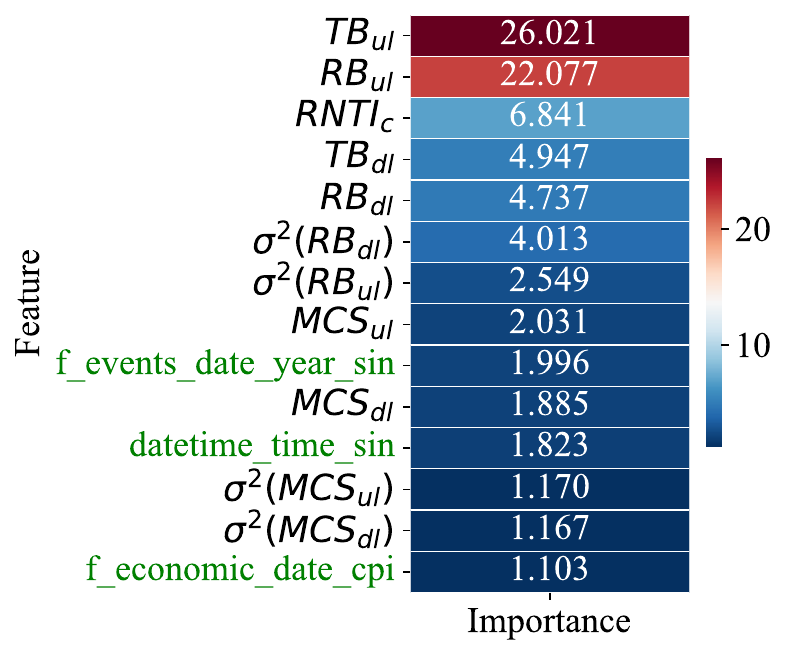}
    \caption{Importance heatmap of original and exogenous features (highlighted in green) added with Upgini.}
    \label{fig:importance_heatmap}
\end{figure}

To conclude the experimental part, we focus on the potential enhancement of our framework's efficacy through the integration of exogenous sources of data into our network data predictors. The main motivation for that is that cellular traffic is heavily affected by exogenous variables, including public holidays, weekends, and specific hours of the day, to mention a few. To capture and leverage the impact of such phenomena in network utilization, we utilized the Upgini Python Library~\cite{Upgini_2022}, which facilitates the search of public datasets to be used in conjunction with our network dataset. 
The Upgini uses an advanced search mechanism that utilizes consecutive experiments with extra ready-to-use ML features from external data sources, such as historical weather data, holidays and events, world economic indicators and markets data, customized for specific locations and dates. Since Upgini enrichment included a significant amount of extra features that generally could not enhance overall performance, a feature selection technique was applied. To find relevant features to enrich our dataset, we measured the importance of various Upgini features by training a gradient-boosting ML model (i.e., Catboost) that includes those features and observing the variation of the model's accuracy in response to feature value changes. 
The process aimed to find out those Upgini-generated features that had similar or even higher importance than the original ones.\footnote{Notice that to fit the considered dataset to Upgini, measurements were aggregated in intervals of 30 minutes by averaging the original data.} The resulting top-14 features are shown in descending order in Fig.~\ref{fig:importance_heatmap}, where the new Upgini features that showed similar or higher importance than the original are highlighted in green. As we can observe, these exogenous features are related to special events (\textit{f\_events\_date\_year\_sin} and \textit{f\_economic\_date\_cpi}) and date time (\textit{datetime\_time\_sin}) and exhibit significant importance within the studied model. Accordingly, we have injected them to our dataset for the next experiment.


\begin{figure}[ht!]
    \centering    
    \includegraphics[width=.9\columnwidth]{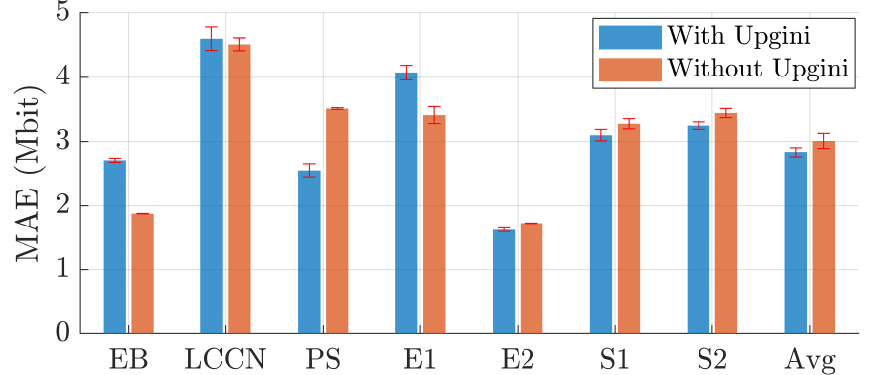}
    \caption{Test MAE across various BS with the presence or absence of exogenous data from Upgini library.}
    \label{fig:upgini_results}
\end{figure}

After identifying the features to be added to our model, we now evaluate the effect of using extra information by training from scratch the LSTM model with the 14 most important features shown in Fig.~\ref{fig:importance_heatmap} under federated settings. The resulting MAE at each BS (with and without Upgini's enhanced dataset) is presented in Fig.~\ref{fig:upgini_results}. As shown, combining exogenous sources of data can have mixed effects on the predictive performance across BS. Specifically,  a decrease in MAE is observed in four out of seven BSs, while the remaining three BSs perform worse when extra features are integrated. On average, the Upgini-enriched dataset leads to 6.55\% lower prediction error, thus offering a promising approach for the task of federated cellular traffic prediction. However, it is important to emphasize the limited generalizability of the introduced features across all datasets. As demonstrated, there was no performance enhancement observed in all BSs, particularly in those experiencing anomalous events (as detailed in Section~\ref{section:dataset}). This observation suggests that feature selection must be conducted carefully, preferably with expert guidance, to improve the efficacy of machine learning models.

\section{Conclusion \& Future Work}
\label{section:conclusions}

This study delves into the rapidly evolving field of mobile traffic forecasting and the integration of FL within the telecommunication industry, with a special focus on the realms of 5G and the upcoming 6G networks. It sheds light on several research directions and identifies key challenges that, once overcome, could substantially improve network efficiency, enhance resource management, and ensure data privacy. The experimental analysis provides useful insights on crucial challenges for the incorporation of FL in mobile traffic forecasting, such as the impact of individual clients, personalization, and integration of data from exogenous data sources. Further, aspects related to FL fine-tuning, including model aggregation and outlier handling, have been evaluated. The presented results demonstrate FL's capability not only in tackling the challenges associated with ML tasks but also in ensuring data privacy and fostering cooperation among network operators sustainably without compromising prediction quality. Overall, the analysis provides a roadmap for advancing mobile traffic forecasting using FL as a privacy-friendly and sustainable solution. By addressing these challenges, the telecommunications industry can expect networks that are more resilient and efficient, paving the way for the next generation of mobile networks. 

However, there are still open problems that need to be addressed in this evolving field. First, explainability in mobile traffic prediction models is becoming increasingly important, especially for network operators and regulators who need to understand the basis of predictions to make well-informed decisions (e.g., related to energy saving). In this respect, techniques such as smooth-graph and layer-wise relevance propagation could provide insight into model decision-making processes, enhancing transparency and trust in predictive models. Second, mobile traffic forecasting can be enhanced by leveraging complementary data sources. A case in point is tempo-spatial correlations, whose comprehension and modeling hold vast potential, especially in densely populated urban areas where user mobility patterns and network usage can vary significantly across short distances. Third, delivering lightweight FL solutions is essential for their viability, and this can be achieved through alternative forms of FL, such as federated distillation and hierarchical FL, and ML model optimization techniques such as model pruning.
Finally, future work could explore the integration of additional factors into the sustainability indicator, such as convergence time and robustness, to provide a more holistic evaluation of learning methods regarding their performance and associated costs.

\section*{Acknowledgment}

This publication has been partially funded by the European Union's Horizon 2020 research and innovation programme under Grant Agreement No. 953775 (GREENEDGE) and the Grant CHIST-ERA-20-SICT-004 (SONATA) by PCI2021-122043-2A/AEI/10.13039/501100011033.
	
\ifCLASSOPTIONcaptionsoff
\newpage
\fi

\bibliographystyle{IEEEtran}
\bibliography{bibliography}

\begin{thebibliography}{10}
\providecommand{\url}[1]{#1}
\csname url@samestyle\endcsname
\providecommand{\newblock}{\relax}
\providecommand{\bibinfo}[2]{#2}
\providecommand{\BIBentrySTDinterwordspacing}{\spaceskip=0pt\relax}
\providecommand{\BIBentryALTinterwordstretchfactor}{4}
\providecommand{\BIBentryALTinterwordspacing}{\spaceskip=\fontdimen2\font plus
\BIBentryALTinterwordstretchfactor\fontdimen3\font minus \fontdimen4\font\relax}
\providecommand{\BIBforeignlanguage}[2]{{%
\expandafter\ifx\csname l@#1\endcsname\relax
\typeout{** WARNING: IEEEtran.bst: No hyphenation pattern has been}%
\typeout{** loaded for the language `#1'. Using the pattern for}%
\typeout{** the default language instead.}%
\else
\language=\csname l@#1\endcsname
\fi
#2}}
\providecommand{\BIBdecl}{\relax}
\BIBdecl

\bibitem{shafiq2011characterizing}
M.~Z. Shafiq \emph{et~al.}, ``{Characterizing and modeling internet traffic dynamics of cellular devices},'' \emph{ACM SIGMETRICS Performance Evaluation Review}, vol.~39, no.~1, pp. 265--276, 2011.

\bibitem{wang2017machine}
M.~Wang \emph{et~al.}, ``{Machine learning for networking: Workflow, advances and opportunities},'' \emph{Ieee Network}, vol.~32, no.~2, pp. 92--99, 2017.

\bibitem{ferriol2022building}
M.~Ferriol-Galm{\'e}s \emph{et~al.}, ``Building a digital twin for network optimization using graph neural networks,'' \emph{Computer Networks}, vol. 217, p. 109329, 2022.

\bibitem{lv2014traffic}
Y.~Lv \emph{et~al.}, ``{Traffic flow prediction with big data: a deep learning approach},'' \emph{IEEE Transactions on Intelligent Transportation Systems}, vol.~16, no.~2, pp. 865--873, 2014.

\bibitem{yin2021deep}
X.~Yin \emph{et~al.}, ``{Deep learning on traffic prediction: Methods, analysis and future directions},'' \emph{IEEE Transactions on Intelligent Transportation Systems}, 2021.

\bibitem{zhang2018long}
C.~Zhang and P.~Patras, ``Long-term mobile traffic forecasting using deep spatio-temporal neural networks,'' in \emph{Proceedings of the Eighteenth ACM International Symposium on Mobile Ad Hoc Networking and Computing}, 2018, pp. 231--240.

\bibitem{cho2015much}
J.~Cho \emph{et~al.}, ``How much data is needed to train a medical image deep learning system to achieve necessary high accuracy?'' \emph{arXiv preprint arXiv:1511.06348}, 2015.

\bibitem{piovesan2021mobile}
N.~Piovesan \emph{et~al.}, ``{Mobile Traffic Forecasting for Green 5G Networks},'' in \emph{2021 IEEE Global Communications Conference (GLOBECOM)}.\hskip 1em plus 0.5em minus 0.4em\relax IEEE, 2021, pp. 1--6.

\bibitem{piovesan2021forecasting}
N.~Piovesan \emph{et~al.}, ``{Forecasting mobile traffic to achieve greener 5G networks: When machine learning is key},'' in \emph{2021 IEEE 22nd International Workshop on Signal Processing Advances in Wireless Communications (SPAWC)}.\hskip 1em plus 0.5em minus 0.4em\relax IEEE, 2021, pp. 276--280.

\bibitem{Ahvar2022}
E.~Ahvar, A.-C. Orgerie, and A.~Lebre, ``Estimating energy consumption of cloud, fog, and edge computing infrastructures,'' \emph{IEEE Transactions on Sustainable Computing}, vol.~7, no.~2, pp. 277--288, 2022.

\bibitem{mcmahan2017communication}
B.~McMahan \emph{et~al.}, ``Communication-efficient learning of deep networks from decentralized data,'' in \emph{Artificial intelligence and statistics}.\hskip 1em plus 0.5em minus 0.4em\relax PMLR, 2017, pp. 1273--1282.

\bibitem{konevcny2016federated}
J.~Kone{\v{c}}n{\`y} \emph{et~al.}, ``{Federated learning: Strategies for improving communication efficiency},'' \emph{arXiv preprint arXiv:1610.05492}, 2016.

\bibitem{perifanis2023federated}
V.~Perifanis \emph{et~al.}, ``{Federated learning for 5G base station traffic forecasting},'' \emph{Computer Networks}, vol. 235, p. 109950, 2023.

\bibitem{trinh2020mobile}
H.~D. Trinh \emph{et~al.}, ``Mobile traffic classification through physical control channel fingerprinting: a deep learning approach,'' \emph{IEEE Transactions on Network and Service Management}, vol.~18, no.~2, pp. 1946--1961, 2020.

\bibitem{box1976time}
G.~E. Box and G.~M. Jenkins, ``{Time series analysis: Forecasting and control San Francisco},'' \emph{Calif: Holden-Day}, 1976.

\bibitem{hill1996neural}
T.~Hill, M.~O'Connor, and W.~Remus, ``Neural network models for time series forecasts,'' \emph{Management science}, vol.~42, no.~7, pp. 1082--1092, 1996.

\bibitem{mikolov2010recurrent}
T.~Mikolov \emph{et~al.}, ``Recurrent neural network based language model,'' in \emph{Interspeech}, vol.~2, no.~3.\hskip 1em plus 0.5em minus 0.4em\relax Makuhari, 2010, pp. 1045--1048.

\bibitem{graves2013speech}
A.~Graves, A.-r. Mohamed, and G.~Hinton, ``Speech recognition with deep recurrent neural networks,'' in \emph{2013 IEEE international conference on acoustics, speech and signal processing}.\hskip 1em plus 0.5em minus 0.4em\relax Ieee, 2013, pp. 6645--6649.

\bibitem{guera2018deepfake}
D.~G{\"u}era and E.~J. Delp, ``Deepfake video detection using recurrent neural networks,'' in \emph{2018 15th IEEE international conference on advanced video and signal based surveillance (AVSS)}.\hskip 1em plus 0.5em minus 0.4em\relax IEEE, 2018, pp. 1--6.

\bibitem{trinh2018mobile}
H.~D. Trinh, L.~Giupponi, and P.~Dini, ``{Mobile traffic prediction from raw data using LSTM networks},'' in \emph{2018 IEEE 29th annual international symposium on personal, indoor and mobile radio communications (PIMRC)}.\hskip 1em plus 0.5em minus 0.4em\relax IEEE, 2018, pp. 1827--1832.

\bibitem{trinh2019urban}
H.~D. Trinh, L.~Giupponi, and P.~Dini, ``{Urban anomaly detection by processing mobile traffic traces with LSTM neural networks},'' in \emph{2019 16th Annual IEEE International Conference on Sensing, Communication, and Networking (SECON)}.\hskip 1em plus 0.5em minus 0.4em\relax IEEE, 2019, pp. 1--8.

\bibitem{qiu2018spatio}
C.~Qiu \emph{et~al.}, ``Spatio-temporal wireless traffic prediction with recurrent neural network,'' \emph{IEEE Wireless Communications Letters}, vol.~7, no.~4, pp. 554--557, 2018.

\bibitem{gu2018recent}
J.~Gu \emph{et~al.}, ``Recent advances in convolutional neural networks,'' \emph{Pattern recognition}, vol.~77, pp. 354--377, 2018.

\bibitem{andreoletti2019network}
D.~Andreoletti \emph{et~al.}, ``Network traffic prediction based on diffusion convolutional recurrent neural networks,'' in \emph{IEEE INFOCOM 2019-IEEE Conference on Computer Communications Workshops (INFOCOM WKSHPS)}.\hskip 1em plus 0.5em minus 0.4em\relax IEEE, 2019, pp. 246--251.

\bibitem{he2016deep}
K.~He \emph{et~al.}, ``{Deep residual learning for image recognition},'' in \emph{Proceedings of the IEEE conference on computer vision and pattern recognition}, 2016, pp. 770--778.

\bibitem{simonyan2014very}
K.~Simonyan and A.~Zisserman, ``Very deep convolutional networks for large-scale image recognition,'' \emph{arXiv preprint arXiv:1409.1556}, 2014.

\bibitem{han2021multivariate}
J.~Han \emph{et~al.}, ``A multivariate-time-series-prediction-based adaptive data transmission period control algorithm for iot networks,'' \emph{IEEE Internet of Things Journal}, vol.~9, no.~1, pp. 419--436, 2021.

\bibitem{perifanis2023towards}
V.~Perifanis \emph{et~al.}, ``Towards energy-aware federated traffic prediction for cellular networks,'' in \emph{2023 Eighth International Conference on Fog and Mobile Edge Computing (FMEC)}.\hskip 1em plus 0.5em minus 0.4em\relax IEEE, 2023, pp. 93--100.

\bibitem{liu2020privacy}
Y.~Liu \emph{et~al.}, ``Privacy-preserving traffic flow prediction: A federated learning approach,'' \emph{IEEE Internet of Things Journal}, vol.~7, no.~8, pp. 7751--7763, 2020.

\bibitem{zeng2021multi}
T.~Zeng \emph{et~al.}, ``Multi-task federated learning for traffic prediction and its application to route planning,'' in \emph{2021 IEEE Intelligent Vehicles Symposium (IV)}.\hskip 1em plus 0.5em minus 0.4em\relax IEEE, 2021, pp. 451--457.

\bibitem{zhang2021dual}
C.~Zhang \emph{et~al.}, ``Dual attention-based federated learning for wireless traffic prediction,'' in \emph{IEEE INFOCOM 2021-IEEE conference on computer communications}.\hskip 1em plus 0.5em minus 0.4em\relax IEEE, 2021, pp. 1--10.

\bibitem{zhang2022efficient}
L.~Zhang, C.~Zhang, and B.~Shihada, ``{Efficient wireless traffic prediction at the edge: A federated meta-learning approach},'' \emph{IEEE Communications Letters}, vol.~26, no.~7, pp. 1573--1577, 2022.

\bibitem{barlacchi2015multi}
G.~Barlacchi \emph{et~al.}, ``{A multi-source dataset of urban life in the city of Milan and the Province of Trentino},'' \emph{Scientific data}, vol.~2, no.~1, pp. 1--15, 2015.

\bibitem{li2020federated}
T.~Li \emph{et~al.}, ``Federated learning: Challenges, methods, and future directions,'' \emph{IEEE signal processing magazine}, vol.~37, no.~3, pp. 50--60, 2020.

\bibitem{huang2023maverick}
J.~Huang \emph{et~al.}, ``Maverick matters: Client contribution and selection in federated learning,'' in \emph{Pacific-Asia Conference on Knowledge Discovery and Data Mining}.\hskip 1em plus 0.5em minus 0.4em\relax Springer, 2023, pp. 269--282.

\bibitem{xue2021toward}
Y.~Xue \emph{et~al.}, ``Toward understanding the influence of individual clients in federated learning,'' in \emph{Proceedings of the AAAI Conference on Artificial Intelligence}, vol.~35, no.~12, 2021, pp. 10\,560--10\,567.

\bibitem{essien2021deep}
A.~Essien \emph{et~al.}, ``A deep-learning model for urban traffic flow prediction with traffic events mined from twitter,'' \emph{World Wide Web}, vol.~24, no.~4, pp. 1345--1368, 2021.

\bibitem{he2020graph}
K.~He \emph{et~al.}, ``Graph attention spatial-temporal network with collaborative global-local learning for citywide mobile traffic prediction,'' \emph{IEEE Transactions on mobile computing}, vol.~21, no.~4, pp. 1244--1256, 2020.

\bibitem{guerra2023cost}
E.~Guerra \emph{et~al.}, ``{The Cost of Training Machine Learning Models over Distributed Data Sources},'' \emph{IEEE Open Journal of the Communications Society}, 2023.

\bibitem{hochreiter1997long}
S.~Hochreiter and J.~Schmidhuber, ``{L}ong short-term memory,'' \emph{{N}eural {C}omputation}, vol.~9, no.~8, 1997.

\bibitem{TS36213}
{3GPP}, \emph{Tech. Specif. Group Radio Access Network; Physical layer procedures (Release 9), {3GPP TS 36.213}}.

\bibitem{anthony2020carbontracker}
L.~F.~W. Anthony, B.~Kanding, and R.~Selvan, ``{Carbontracker: Tracking and predicting the carbon footprint of training deep learning models},'' \emph{arXiv preprint arXiv:2007.03051}, 2020.

\bibitem{liu2008isolation}
F.~T. Liu, K.~M. Ting, and Z.-H. Zhou, ``Isolation forest,'' in \emph{2008 eighth ieee international conference on data mining}.\hskip 1em plus 0.5em minus 0.4em\relax IEEE, 2008, pp. 413--422.

\bibitem{rousseeuw1993iqr}
P.~J. Rousseeuw and C.~Croux, ``Alternatives to the median absolute deviation,'' \emph{Journal of the American Statistical association}, vol.~88, no. 424, pp. 1273--1283, 1993.

\bibitem{jordaan2004svm}
E.~M. Jordaan and G.~F. Smits, ``Robust outlier detection using svm regression,'' in \emph{IEEE International Joint Conference on Neural Networks}, vol.~3, 2004, pp. 2017--2022.

\bibitem{blazquez2021review}
A.~Bl{\'a}zquez-Garc{\'\i}a \emph{et~al.}, ``A review on outlier/anomaly detection in time series data,'' \emph{ACM computing surveys (CSUR)}, vol.~54, no.~3, pp. 1--33, 2021.

\bibitem{wang2020fednova}
J.~Wang \emph{et~al.}, ``Tackling the objective inconsistency problem in heterogeneous federated optimization,'' \emph{Advances in neural information processing systems}, vol.~33, pp. 7611--7623, 2020.

\bibitem{tzu2020fedavgm}
T.-M.~H. Hsu, H.~Qi, and M.~Brown, ``Federated visual classification with real-world data distribution,'' in \emph{Computer Vision--ECCV 2020: 16th European Conference, Glasgow, UK, August 23--28, 2020, Proceedings, Part X 16}.\hskip 1em plus 0.5em minus 0.4em\relax Springer, 2020, pp. 76--92.

\bibitem{reddi2020adaptive}
S.~Reddi \emph{et~al.}, ``Adaptive federated optimization,'' \emph{arXiv preprint arXiv:2003.00295}, 2020.

\bibitem{jiang2023personalisation}
Y.~Jiang, S.~Chen, and X.~Bao, ``Amplitude-aligned personalization and robust aggregation for federated learning,'' \emph{IEEE Transactions on Sustainable Computing}, 2023.

\bibitem{Upgini_2022}
\BIBentryALTinterwordspacing
R.~Post and N.~Toro, ``{Upgini - automated data search \& enrichment library},'' Jul. 2022. [Online]. Available: \url{https://github.com/upgini/upgini}
\BIBentrySTDinterwordspacing

\end{thebibliography}

\begin{IEEEbiography}[{\includegraphics[width=1in,height=1.25in,clip,keepaspectratio]{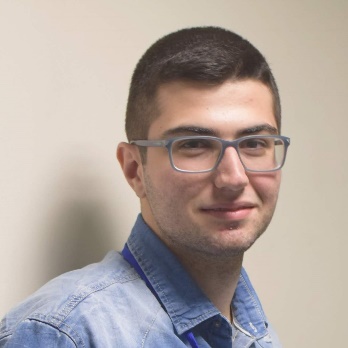}}]{Nikolaos Pavlidis} is a PhD student at the Dept. of Electrical and Computer Engineering of the Democritus University of Thrace (Greece) and holds a BSc from the same department with honors. His main research interests are: federated learning, IoT, carbon aware computations and graph-based machine learning algorithms.
\end{IEEEbiography}

\begin{IEEEbiography}[{\includegraphics[width=1in,height=1.25in,clip,keepaspectratio]{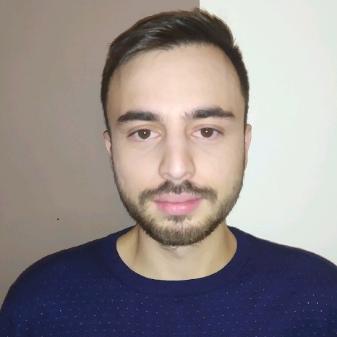}}]{Vasileios Perifanis} is a Post-doc Researcher. He received his PhD from the Dept. of Electrical and Computer Engineering of the Democritus University of Thrace (Greece). He has a strong background in machine learning, with a focus on privacy-preserving federated learning, distributed and efficient computations.
\end{IEEEbiography}

\begin{IEEEbiography}[{\includegraphics[width=1in,height=1.25in,clip,keepaspectratio]{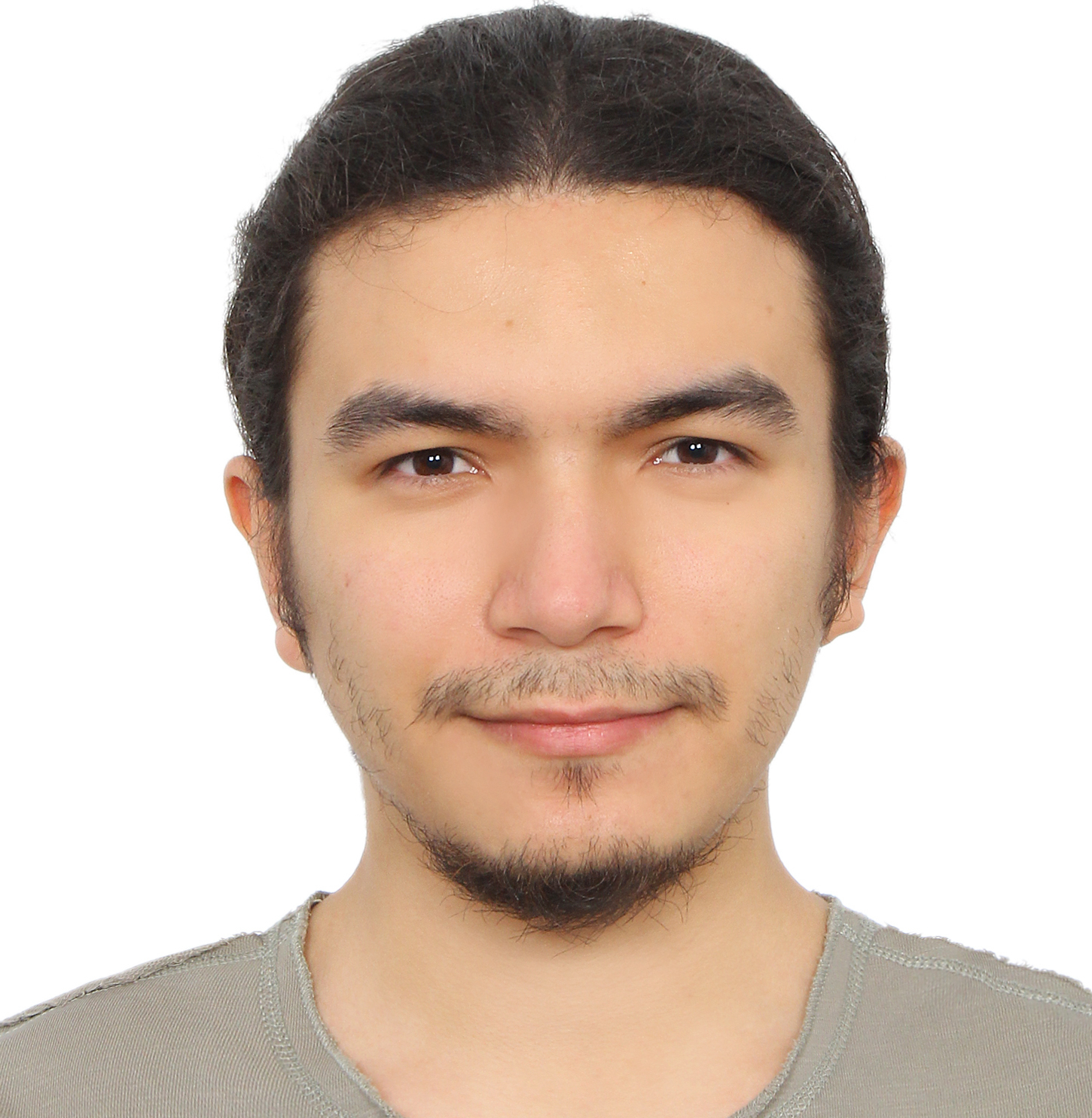}}]{Selim F. Yilmaz} is a PhD student at the Dept. of Electrical and Electronic Engineering of Imperial College London (United Kingdom). He received his B.S. in computer engineering and M.S. degree in electrical and electronic engineering from Bilkent University (Türkiye), in 2019 and 2021, respectively. His main research interests include edge learning and inference, and source-channel coding.
\end{IEEEbiography}

\begin{IEEEbiography}[{\includegraphics[width=1in,height=1.25in,clip,keepaspectratio]{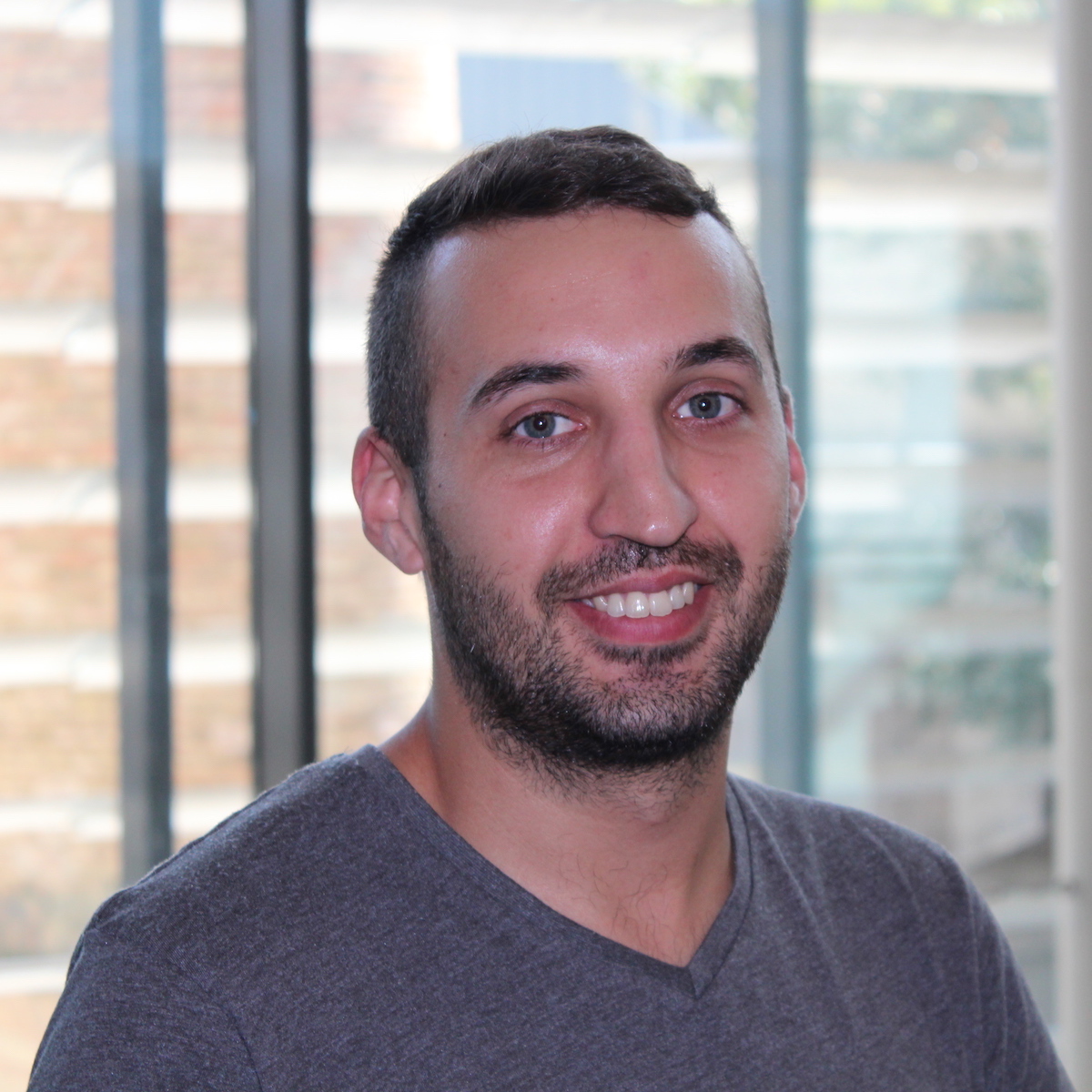}}]{Francesc Wilhelmi} is a research engineer at Nokia Bell Labs. He holds a Ph.D. in Information and Communication Technologies (2020) and an M.Sc. in Intelligent and Interactive Systems (2016) from Universitat Pompeu Fabra (UPF). His main research interests are: Wi-Fi and its evolution, network simulators and network digital twinning, machine learning, and distributed ledger technology.
\end{IEEEbiography}

\begin{IEEEbiography}[{\includegraphics[width=1in,height=1.25in,clip,keepaspectratio]{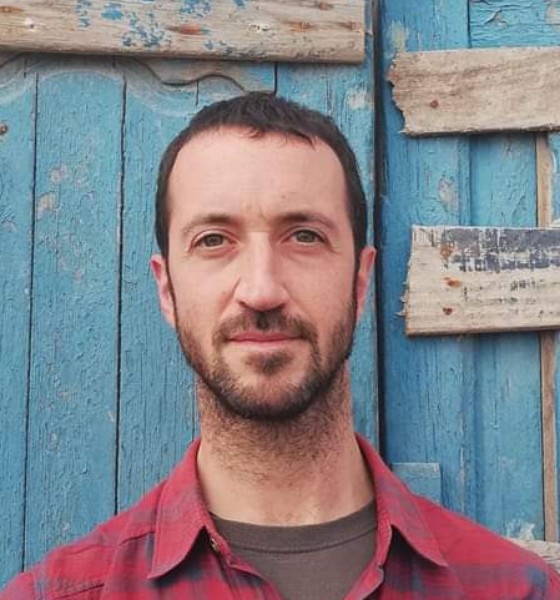}}]{Marco Miozzo} received his M.Sc. degree in Telecommunication Engineering from the University of Ferrara (Italy) in 2005 and the Ph.D. from the Technical University of Catalonia (UPC) in 2018. In June 2008 he joined the Centre Tecnologic de Telecomunicacions de Catalunya (CTTC).
His main research interests are: sustainable mobile networks, green wireless networking, energy harvesting, multi-agent systems, machine learning, green AI, explainable AI.
\end{IEEEbiography}

\begin{IEEEbiography}[{\includegraphics[width=1in,height=1.25in,clip,keepaspectratio]{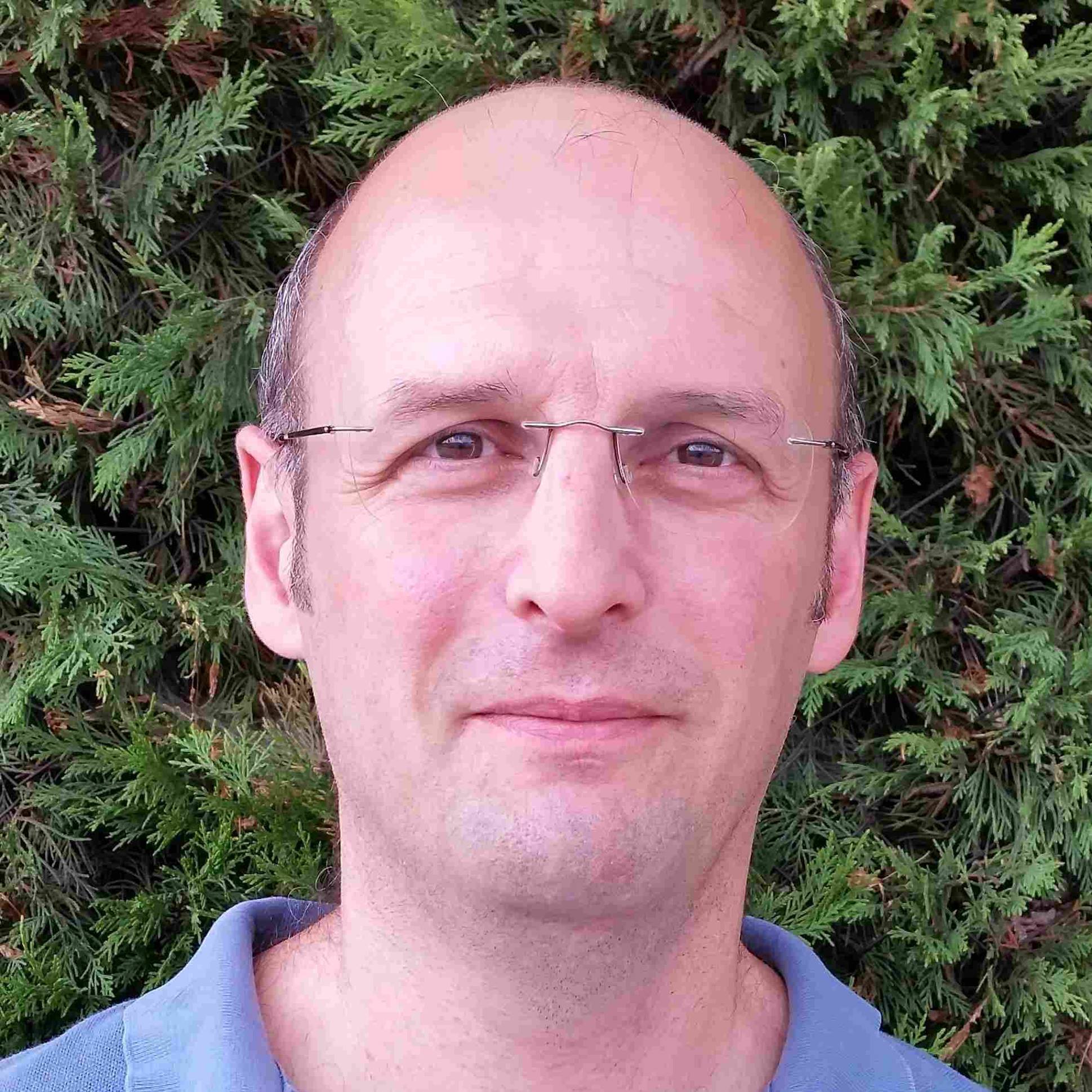}}]{Pavlos S. Efraimidis} is a Professor of Computer Science at the Dept. of Electrical and Computer Engineering of the Democritus University of Thrace (Greece) and an Adjunct Researcher of the Athena Research Center. He received his PhD in Informatics in 2000 from the Univ. of Patras and the diploma of Computer Engineering and Informatics from the same university in 1995. His current research interests are in the fields of design and analysis of algorithms, graph theory and network analysis, federated learning, and algorithmic aspects of privacy.
\end{IEEEbiography}

\begin{IEEEbiography}[{\includegraphics[width=1in,height=1.25in,clip,keepaspectratio]{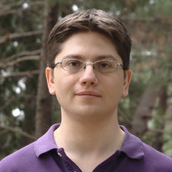}}]{Remous-Aris Koutsiamanis} is an Associate Professor at IMT Atlantique (Nantes, France) and a member of the STACK team of LS2N, a joint team with INRIA, focusing on the networking aspects of geo-distributed edge systems.
He received his PhD from the Department of Electrical and Computer Engineering of the Democritus University of Thrace, Greece in February 2016 on the application of game theory to network QoS problems. His main research interests are in IoT, QoS and distributed resource management.
\end{IEEEbiography}

\begin{IEEEbiography}[{\includegraphics[width=1in,height=1.25in,clip,keepaspectratio]{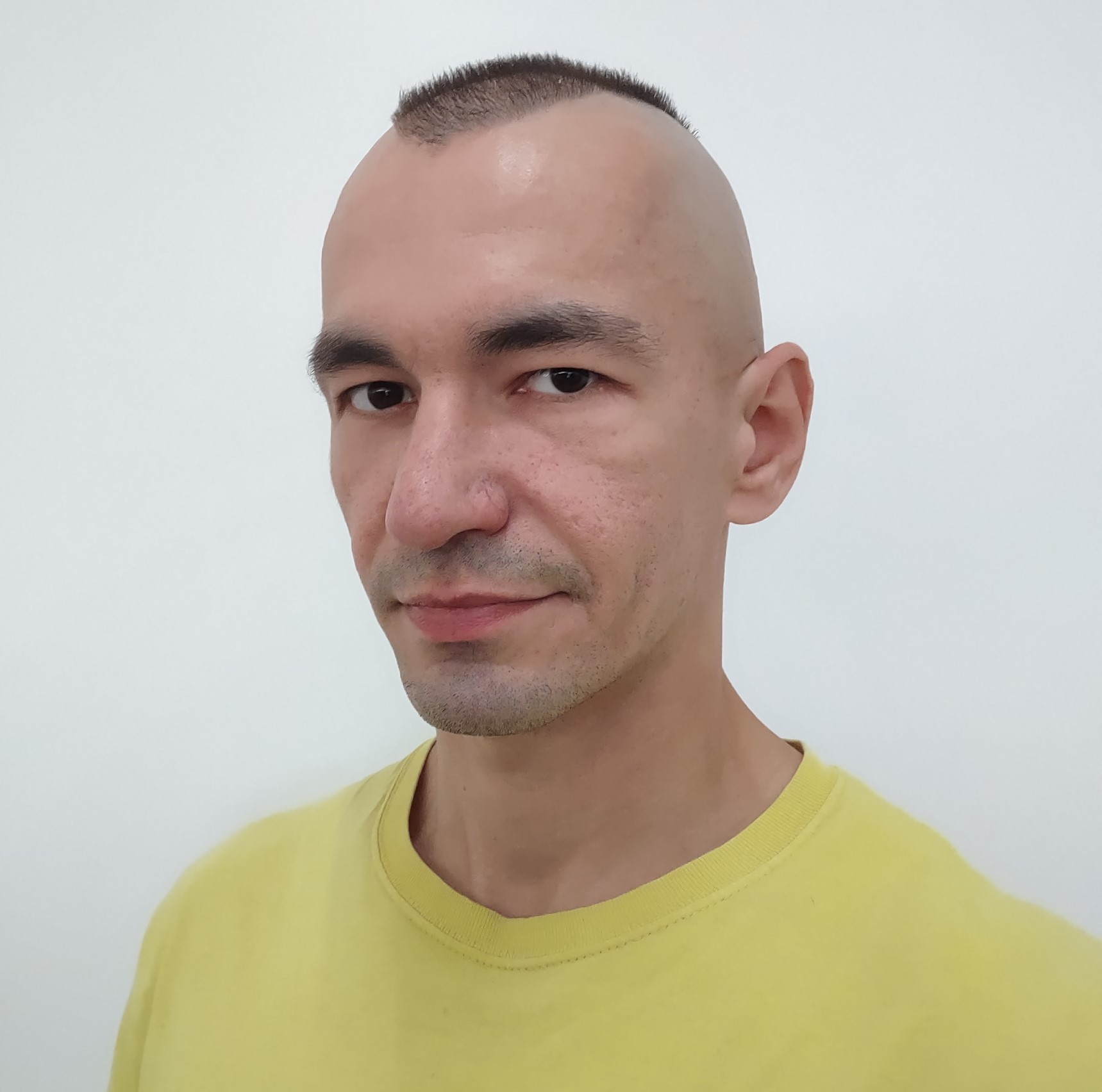}}]{Pavol Mulinka} is researcher at Centre Tecnologic de Telecomunicacion de Catalunya (CTTC) now working on a SUCCESS-6G project. He received M.Eng. at Faculty of Electrical Engineering and Information Technology, Slovak University of Technology in Bratislava, Telecommunications department, and the Ph.D. at the Faculty of Electrical Engineering, Czech Technical University in Prague, Telecommunications department. He is also an active data science volunteer on GitHub, mainly contributing to pytorch-widedeep library. He is a machine learning enthusiast, data scientist, Python programmer, and former network engineer.
\end{IEEEbiography}

\begin{IEEEbiography}[{\includegraphics[width=1in,height=1.25in,clip,keepaspectratio]{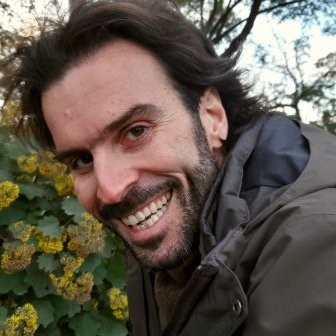}}]{Paolo Dini} is leading researcher at Centre Tecnologic de Telecomunicacion de Catalunya (CTTC), where coordinates the activities of the Sustainable Artificial Intelligence research unit. He received PhD in Information and Communication Technology and laurea in Electronic Engineering in in 2005 and 2001, respectively, by the University of Rome "La Sapienza". His research interests encompass machine learning, optimal control, sustainable computing, multi-agent systems, cyber-physical systems, energy efficiency. 
\end{IEEEbiography}

\vspace{11pt}

\vfill

\end{document}